%% file: main.tex
\documentclass[journal]{IEEEtran}
\usepackage{times}
\usepackage{latexsym}
\usepackage{graphicx}
\usepackage{subfigure}
\usepackage{url}
\usepackage{color}
\usepackage[linesnumbered, ruled]{algorithm2e}

\newcommand{\rev}[1]{\textcolor{black}{#1}}

\newcommand{\minrev}[1]{\textcolor{black}{#1}}

\definecolor{mygray}{gray}{0.7}
\usepackage{multirow}
\usepackage{booktabs}

\ifCLASSINFOpdf
\else
\fi
\usepackage{amsmath}
\usepackage{amssymb}
\def\Bo{\mbox{\boldmath $o$}}

\def\Bw{\mbox{\boldmath $w$}}
\def\Be{\mbox{\boldmath $e$}}

\def\Bd{\mbox{\boldmath $d$}}

\def\Bp{\mbox{\boldmath $p$}}
\def\Bc{\mbox{\boldmath $c$}}

\def\Bc{\mbox{\boldmath $c$}}

\def\Bw{\mbox{\boldmath $w$}}

\def\Bo{\mbox{\boldmath $o$}}

\def\Bo{\mbox{\boldmath $o$}}

\begin{document}
%
\title{Learning in Text Streams: Discovery and Disambiguation of Entity and Relation Instances\thanks{\textcopyright 2020 IEEE. Personal use of this material is permitted. Permission from IEEE must be obtained for all other uses, in any current or future media, including reprinting/republishing this material for advertising or promotional purposes, creating new collective works, for resale or redistribution to servers or lists, or reuse of any copyrighted component of this work in other works.}}
%
%
%

\author{Marco~Maggini, Giuseppe~Marra, Stefano~Melacci, \IEEEmembership{Member,~IEEE}, Andrea~Zugarini
\thanks{M. Maggini and S. Melacci are with the Department of Information Engineering and Mathematics, University of Siena, Siena, Italy. G. Marra and A. Zugarini are with the University of Florence, Florence, Italy.}

}

%
%

\markboth{Now Published in: IEEE Transactions on Neural Networks and Learning Systems. DOI: \texttt{10.1109/TNNLS.2019.2955597}}%
{Shell \MakeLowercase{\textit{et al.}}: Bare Demo of IEEEtran.cls for IEEE Journals}
%



\maketitle

\begin{abstract}
	\input{abstract.tex}
\end{abstract}

\section{Introduction}
\label{sec:intro}
\input{introduction.tex}

\section{Problem Setting}
\label{sec:problem}
\input{problem.tex}

\section{Model}
\label{sec:model}
\input{model.tex}

\section{\rev{Online Learning Dynamics}}
\label{sec:learning}
\input{learning.tex}

\section{Related Work}
\label{sec:related}
\input{related.tex}

\section{Experiments}
\label{sec:exp}
\input{exp.tex}

\section{Conclusions and Future Work}
\label{sec:future}
\input{conclusions.tex}





%
\bibliography{tnnls}
\bibliographystyle{IEEEtran}

%

\begin{IEEEbiography}[{\includegraphics[width=1.2in,height=1.25in,clip,keepaspectratio]{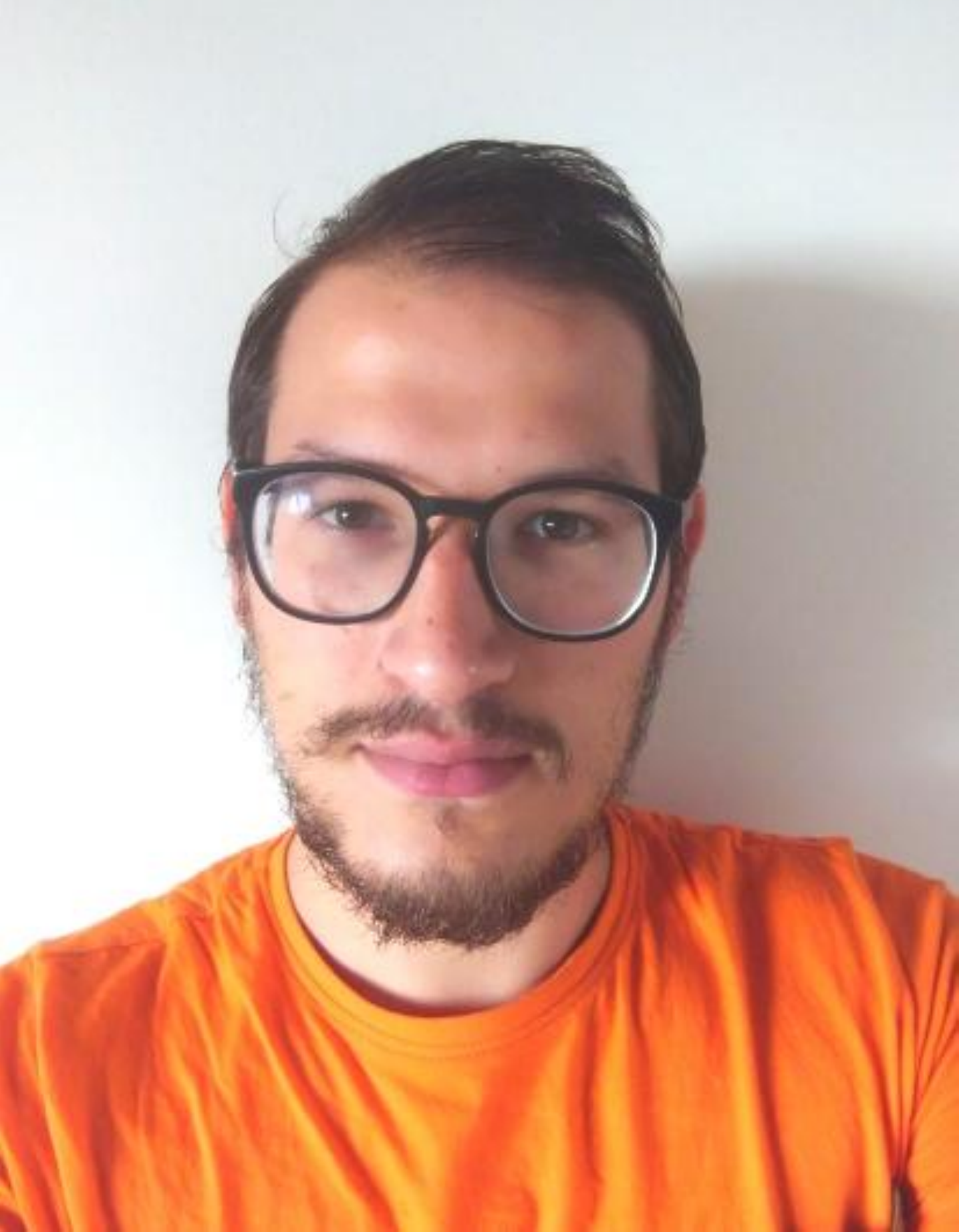}}]{Giuseppe Marra} received the M.S. Degree in ICT Engineering (cum laude) in 2014. Currently, he is a PhD student of Smart Computing program at Universities of Florence and Siena and a visiting scholar at Katholieke Universiteit Leuven. His research interests include the integration of logical reasoning and deep learning, particularly focused on Natural Language Processing applications. He worked on applied Web Semantics in the R\&D Lab of Engineering Tributi S.p.A, Trento. 
\end{IEEEbiography}

\begin{IEEEbiography}[{\includegraphics[width=1.2in,height=1.25in,clip,keepaspectratio]{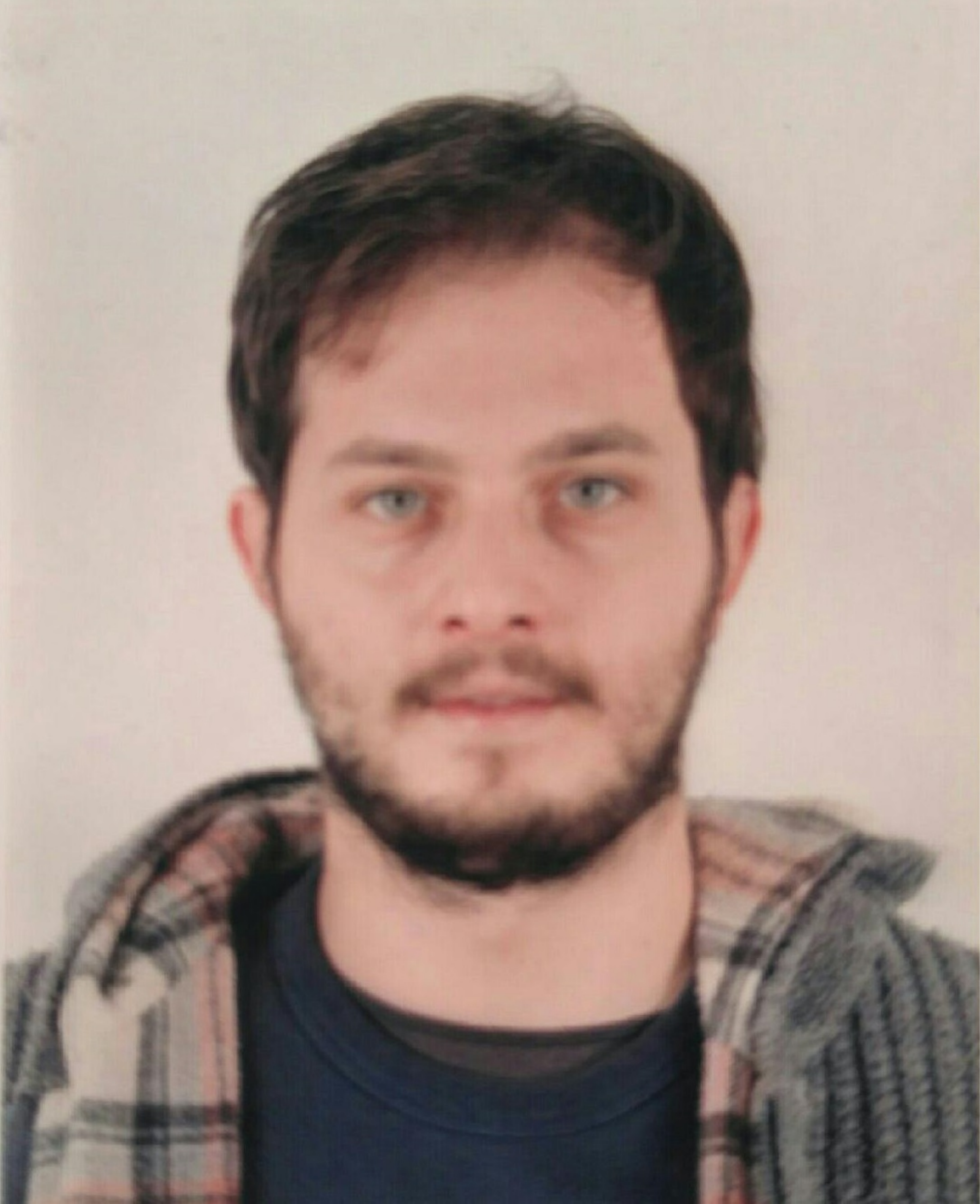}}]{Andrea Zugarini}
received the M.S. Degree in Computer and Automation Engineering (cum laude) in 2017. Currently, he is a PhD student of Smart Computing program at Universities of Florence and Siena. He is interested in Machine-Learning applications to Natural Language Processing with emphasis on Conversational Agents. 
\end{IEEEbiography}

\begin{IEEEbiography}[{\includegraphics[width=1in,height=1.2in,clip,keepaspectratio]{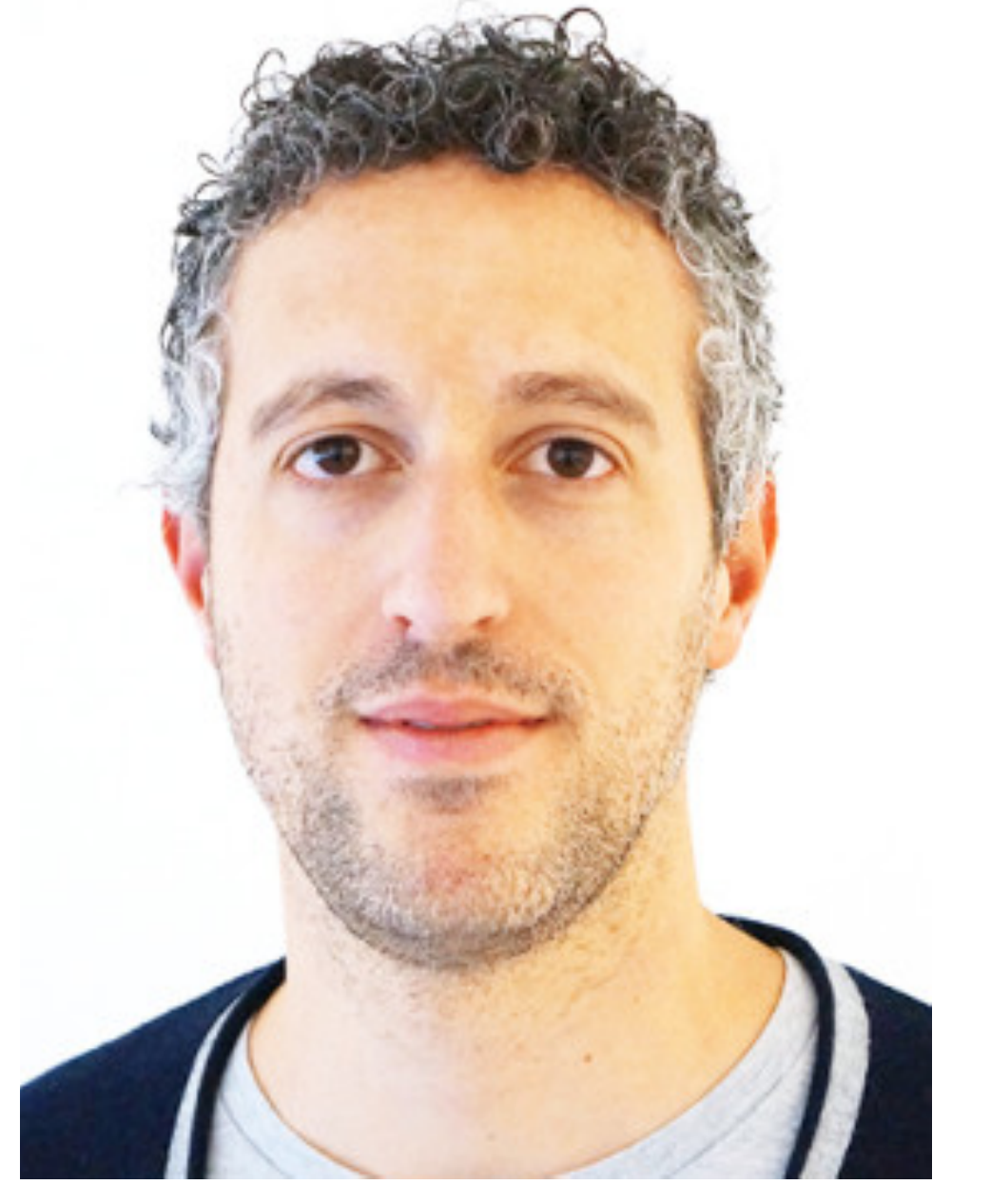}}]{Stefano Melacci}
received the M.S. Degree in Computer Engineering (cum laude) and the PhD degree in Computer Science (Information Engineering) from the University of Siena, Italy, in 2006 and 2010, respectively. He worked as Visiting Scientist at the Computer Science and Engineering Department of the Ohio State University, Columbus, USA, and he is currently Assistant Professor of the Department of Information Engineering and Mathematics, University of Siena. His research interests include machine learning and pattern recognition, mainly focused on Neural Networks and Kernel Machines, with applications to Computer Vision and Natural Language Processing. He serves as Associate Editor of the IEEE Transactions on Neural Network and Learning Systems.
\end{IEEEbiography}

\begin{IEEEbiography}[{\includegraphics[width=1in,height=1.2in,clip,keepaspectratio]{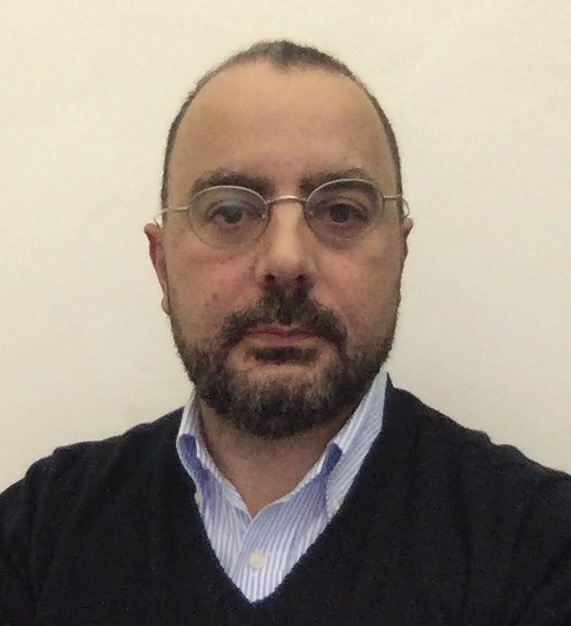}}]{Marco Maggini}
received the M.S. degree (cum laude) in Electronic Engineering in February 1991, and the PhD in Computer Science and Control Systems in 1995 from the University of Firenze. He is currently full professor in Computer Engineering at the Department of Information Engineering and Mathematical Sciences of the University of Siena. His main research interests are on machine learning, with focus on neural networks and kernel methods, and applications in Information Retrieval, Information Extraction, Natural Language Processing, and Pattern Recognition. He is coauthor of more than one hundred scientific publications. He has served as an associate editor of the ACM Transactions on Internet Technology and he has been member of the program committees of several international conferences.
\end{IEEEbiography}




\end{document}

%% file: abstract.tex
We consider a scenario where an artificial agent is reading a stream of text composed of a set of narrations, and it is informed about the identity of some of the individuals \rev{that are mentioned in the  text portion that is currently being read}. The agent is expected to learn to follow the narrations, thus disambiguating mentions and discovering new individuals. 
We focus on the case in which individuals are entities and relations, and we propose an end-to-end trainable memory network that learns to discover and disambiguate them in an online manner, performing one-shot learning, and dealing with a small number of sparse supervisions. Our system builds a not-given-in-advance knowledge base, and it improves its skills while reading unsupervised text. The model deals with abrupt changes in the narration, taking into account their effects when resolving co-references. We showcase the strong disambiguation and discovery skills of our model on a corpus of Wikipedia documents and on a newly introduced dataset, that we make publicly available.

%% file: introduction.tex

Most of nowadays machine-learning-based approaches to language-related problems are defined in the classical setting in which we exploit a batch of data to build powerful, offline predictors, especially when the data are paired with supervisions. Little has been done when considering the setting in which we process a continuous stream of text and build systems that learn and respond in an online manner.
However, conversational systems \cite{yu2016strategy}, information extractors from streams of news \cite{del2005ranking}, or social network data \cite{ritter2012open}, and those systems that require interactions with the environment \cite{christakopoulou2016towards}, belong to the latter setting. Supervisions are usually sparse and very limited, and the system is expected to perform ``one-shot'' learning and to quickly react to them. 

\rev{Our work faces the problem of learning to extract information while reading a text stream, with the aim of identifying entities and relations in the text portion that is currently being read. This problem is commonly tackled by assuming the existence of a Knowledge Base (KB) of entities and relations, where several entity/relation instances are paired with additional information, such as the common ways of referring to them or sentences/facts in which they are involved. 
Then, once an input sentence is provided for reading, sub-portions of text must be linked to entity or relation instances of the KB. The linking process introduces the challenging issue of dealing with multiple distinct entities (relations) that are mentioned with the same text, and thus the system has to disambiguate which is the ``right'' entity or relation instance of the KB for the considered text fragment. In particular, the context around the fragment or, if needed, information that was provided in the previous sentences of the text stream can be used to perform the disambiguation.
As a very simple example, consider the sentence \textit{Clyde went to the office}, being \textit{Clyde} and \textit{the office} two text fragments that indicate entities, while \textit{went to} is text that is about a relation. \textit{Clyde} could be the mention that is used to indicate different people in the KB, and several offices could be mentioned by the expression \textit{the office} (mentions to relations follow the same logic).}

At a first glance, this problem shares basic principles and intuitions with several existing methods, such as Entity Linking \cite{shen2015entity}, Word Sense Disambiguation \cite{raganato2017word}, Named Entity Recognition \cite{chiu2016named}, Knowledge Population \cite{rajani2016combining}, and others (we postpone to Section \ref{sec:related} the description of the related approaches). 
\rev{However, the problem we consider in this paper is way more challenging, since the aforementioned KB is not given in advance, and the system has to take care of progressively building it, while also using such KB to disambiguate the input data. This means that the system is required to decide whether a certain mention is about an entity/relation instance already inserted into the KB or if it is about a never-seen-before entity/relation, and, in that case, to update the KB. Moreover, since we deal with streams of text, learning is performed in an online manner, and the system has to take a decision before processing the following sentence. }

\rev{ Motivated by the aforementioned challenges, in this paper we propose an end-to-end memory-augmented trainable system that learns to discover and disambiguate entity/relation instances in a text stream, exploiting a small number of sparse supervisions\footnote{\rev{Supervisions are indications of the precise entity/relation instance that a fragment of text refers to.}} and operating in an online manner}.
\rev{In particular, this work makes the following contributions. (1) We propose a new online-learning method for populating a not-given-in-advance KB of entities and relations. (2) We introduce a new scheme to learn latent representations of entities and relations directly from data, either autonomously (self-learning) or using limited supervisions. Character-based encoders are exploited to handle small morphological variations (plurals, suffixes, ...) and typos, synonymy, semantic similarity. (3) We present a new problem setting where the system is evaluated while reading a stream of sentences organized into short stories, requiring online learning capabilities. (4) We showcase the validity of our approach both in an existing dataset made of Wikipedia articles and a new dataset that we introduce and make publicly available.}


This paper is organized as follows. Section \ref{sec:problem} formally describes the problem we face. 
 \rev{The inferential process of the proposed architecture is described in Section \ref{sec:model}, while online learning dynamics are described in Section \ref{sec:learning}}. Related work is reviewed in Section \ref{sec:related}. Experiments and future plans are reported in Section \ref{sec:exp} and \ref{sec:future}, respectively.


%% file: problem.tex

We consider a continuous stream of text, that at each time step $t$ produces a sentence $s_t$. Groups of contiguous sentences are organized into small \textit{stories} about a (not-known-in-advance) set of actors/objects, so that the narration is discontinuous whenever a new story begins. 

\begin{figure}[!ht]
\hskip -2mm
\includegraphics[width=0.5\textwidth]{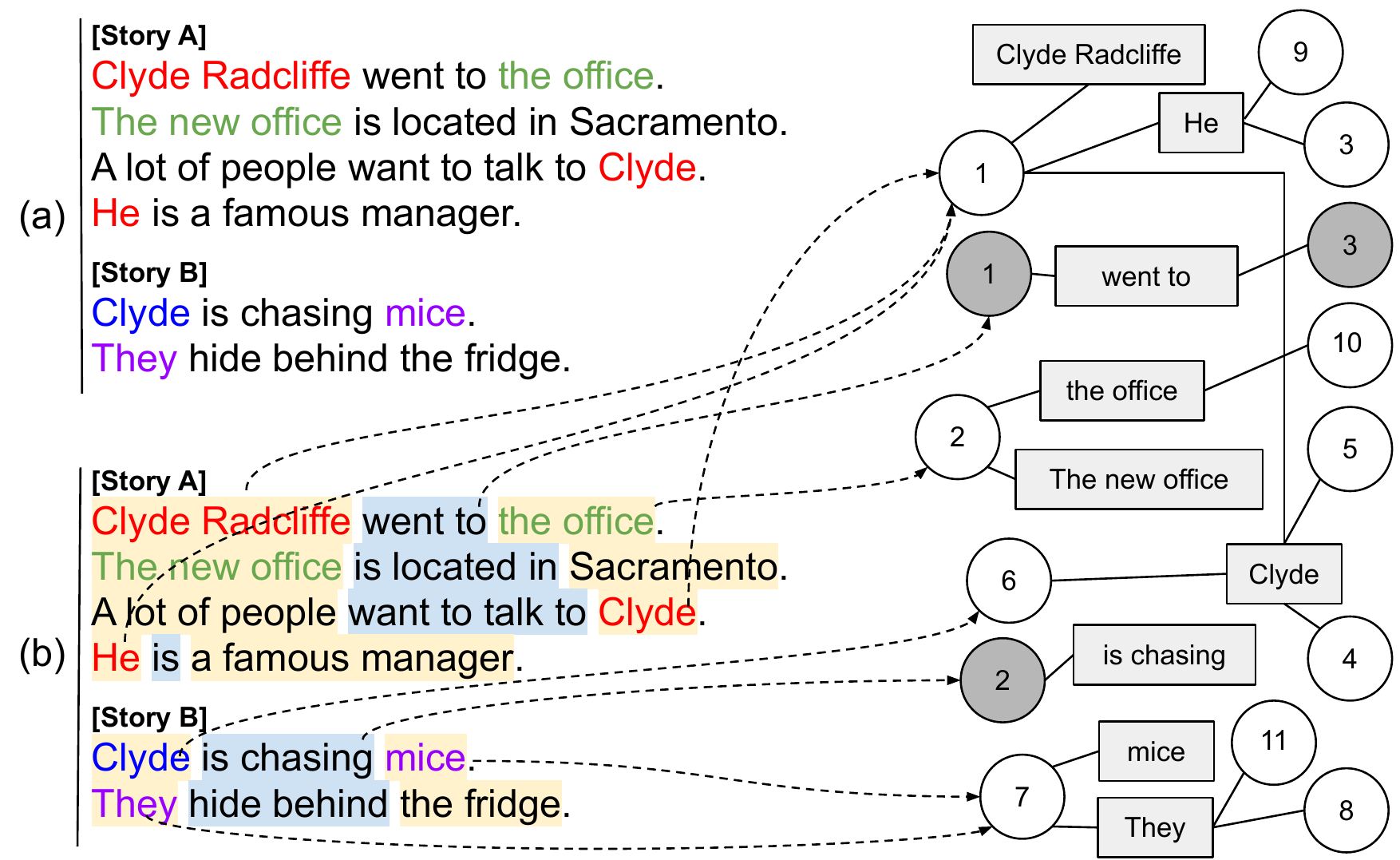}
\caption{\rev{Left: text stream composed of two stories. Right: an example of KB.} (a) Input: sentences from the stream. (b) System output: mentions to entities and relations are detected (pale yellow and pale blue background, respectively), and linked (dashed lines) to KB \rev{instances} (circles) - only some links are shown, for clarity. 
\rev{Empty circles are entity instances, while filled-grey circles are relation instances (circles are intended to also include context-related information that characterize the instances). Boxes represent known mentions, and they are connected with the compatible instances.} \textit{We printed with the same color those mentions that should be linked to the same instance}.}
\label{fig:ex}
\end{figure}

We focus on the problem of developing a system that, given $s_t$, produces its interpretation by linking text fragments to a KB, that the system is responsible of creating and updating (Figure \ref{fig:ex}). \rev{We think of the KB as a set of \textit{instances}, and the considered text fragments of $s_t$ are \textit{mentions} to them. Some mentions are about \textit{entities}, others are about \textit{relations}. For each instance, the KB stores (possibly) multiple mentions that are commonly used to refer to the instance itself, as for entity $1$ and $2$ in Figure \ref{fig:ex}}. On the other way around, the same mention can be shared by more than one instance, as in the example of Figure \ref{fig:ex} where the same textual form \textit{Clyde} refers to different entities (entities $1$ and $6$).
\rev{KB instances also include information about the \textit{contexts} in which instances have been mentioned in the stream so far, where the notion of \textit{context} compactly indicates the whole sentence in which the instance was mentioned. Here and throughout the paper, we simplify the descriptions by frequently using the generic term \textit{instance} to refer to both the cases of relations and entities, without making a precise distinction (if not needed).
}

\rev{We consider the case in which text fragments of $s_t$ are matched with the mentions in the KB to detect compatible instances. Then, instances are disambiguated by observing the current context (that is the portion of $s_t$ around the mention), and exploiting the knowledge about the story to which $s_t$ belongs (up to the current sentence)}. 
At the beginning of each story, a small number of sentences are supervised with the identity of the mentioned entities and relations.
The system is expected to follow the narration by disambiguating mentions, learning from such sparse supervisions (and quickly reacting to them), and discovering new instances.
The natural ambiguity of language, the discontinuous narration, and the dynamic nature of the KB, require the system to develop advanced disambiguation procedures to interpret $s_t$.

\rev{It is worth noticing that, in the proposed setting, the system decisions on a given sentence are evaluated immediately when processing the following sentences, regardless of whether the sentence was supervised or not. This schema gives rise to a novel online dynamics of the learning process, which is profoundly different from most of the existing approaches, being it more challenging and realistic than common batch-mode offline approaches (consider, for example, the case in which it is framed within conversational applications). This work encourages the development of new methods,  models and datasets, for this extremely relevant, but largely unexplored setting.}

\rev{The proposed model has the potentiality of being enhanced by introducing a more structured and advanced knowledge component (types, facts, abstract notions for higher level inference - such as logic formulas). For example, once entities and relations are detected from a sentence $s_t$, they could be combined together to create symbolic representations of facts. Thus the KB could be represented as a knowledge graph, where entities are vertices and relations are the edges of the graph. As long as the knowledge base becomes more structured, the added information may be used to enhance the disambiguation procedure, or to introduce a further level of abstraction on which symbolic reasoning can be performed.
In this work we do not face the problem of designing an enhanced knowledge base module, but we focus on the tasks of instance discovery and disambiguation in text streams, that provide the basic functionalities required to link the input text to the instances in the KB.}

\rev{Section \ref{sec:model} will describe our model from an operational point of view, while Section \ref{sec:learning} will describe the learning dynamics associated to the online learning framework}.

%% file: model.tex
At each time step, the system processes an input sentence $s$ (we drop the time index, for simplicity), and it \rev{ detects those text portions $z$ (i.e. one or more adjacent tokens) that are expected to mention KB instances. This task is referred to as {\em mention detection}.
For each candidate mention $z$ in $s$, the system predicts how much each KB instance is compatible with $z$ in the context where it appears in $s$. The prediction scores are collected in a vector $\Bo$, where each entry corresponds to a given instance in the KB. Then, then mention is linked to the most-likely instance $y$ as }
\begin{equation}
\Bo = \phi(z,s), \quad y = \arg\max(\Bo) \ ,
\label{ooo}
\end{equation}
\rev{ where $\phi$ is the function that computes the affinity scores of $z$ with respect to all the KB instances, given the context of the current sentence $s$. In this work, we replace the second argument of $\phi$ by the sequence of mentions detected in $s$, excluding $z$ itself, i.e., $\phi(z, \{z_j : z_j \in s,  z_j \neq z\})$.}

\begin{figure}[!ht]
\includegraphics[width=0.48\textwidth, trim=0 1cm 0 0]{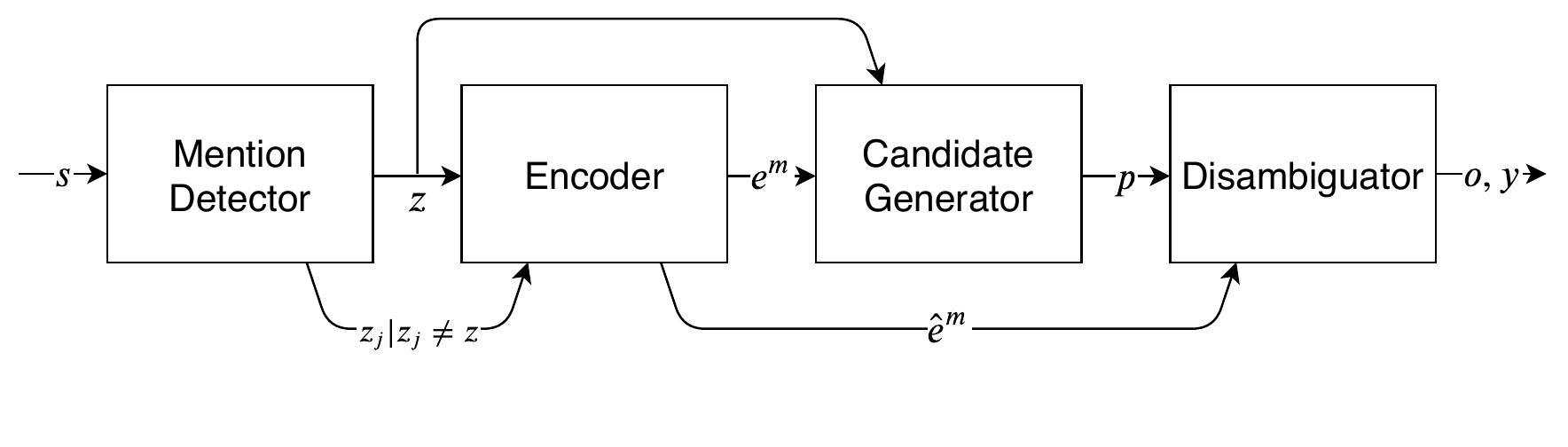}
\caption{\rev{Computational flow of the model for mention-instance linking. The system can be seen as a sequential composition of sub-systems that process the input sentence $s$ and finally output the identifiers of the instances that are linked to the mentions detected in $s$ ($z$ is a generic mention detected in $s$).} }
\label{fig:flow_chart}
\end{figure}

\rev{The whole system is the composition of multiple computational modules, that we sketch in Figure \ref{fig:flow_chart}.
The \textsc{Mention Detector} segments $s$ by identifying mentions to entities and relations. For example, \textit{Parry is chasing a mouse} is segmented into two mentions to entities (i.e ``Parry'' and ``a mouse'') and a mention to a relation (i.e ``is chasing''). Each mention $z$ and its context are encoded into vectorial representations (i.e. embeddings) by the \textsc{Encoder} module. The \textsc{Candidate Generator} produces a probability distribution over the KB instances, by combining different sources of information (e.g. surface form of the mention, embedding, temporal coherence in the current story). Finally, a \textsc{Disambiguator} takes the final decision on which is the most likely KB instance to link, using the aforementioned probability distribution and the embedding of the mention context.}
\rev{ Section 
\ref{sec:seg}, \ref{sec:enc}, \ref{sec:cg}, \ref{sec:du} describe each  computational block in details. 
}

\subsection{Mention Detection}
\label{sec:seg}
\rev{The goal of the \textsc{Mention Detector} (first block of the pipeline in Figure~\ref{fig:flow_chart}) is to segment each sentence into non-overlapping text fragments, which are mentions to yet-unknown entity or relation instances.}

\rev{Motivated by the need of developing models that are robust to morphological changes and that do not depend on a pre-defined vocabulary of words (as needed by interactive/conversational applications), we process the input data using a character-level encoding, following the approach we proposed in \cite{marra2018unsup}. In particular, Bidirectional Recurrent Neural Networks (BiRNN) are exploited to build vectorial representations of words at multiple levels. Given an input sentence $s$, composed of $v$ words $x_i$, $i \in [1,v]$, the \textit{word embedding} $\Be_i^{w}$ of the word $x_i$ is
\begin{equation}
	\label{eq:WE}
	\Be_i^{w} = \overleftrightarrow{r_{c,w}}(\hat{\Bc}_{i,1} \ldots \hat{\Bc}_{i,|x_i|}) \ ,
\end{equation}
where $\hat{\Bc}_{i,k}$ is the $k$-th character of the $i$-th word and $\overleftrightarrow{r_{c,w}}$ is a BiRNN outputting the concatenation of its hidden states in both directions.}
\rev{Since the morphological representation of a word is usually meaningless if taken isolated from its context, we compute the \textit{contextualized word embedding} $\tilde \Be_i^{w}$ of the word $x_i$ as
\begin{equation}
\label{eq:CWE}
	\tilde{\Be}_i^{w}  = [\overrightarrow{{r}_{e,w}}(\Be_{1}^{w} \ldots, \Be_{i}^{w}), \overleftarrow{{r}_{e,w}}(\Be_{n}^{w}\ldots \Be_{i}^{w})] \ ,
\end{equation} 
where $\overrightarrow{{r}_{e,w}}$ and $\overleftarrow{{r}_{e,w}}$ are two RNNs processing the character-level embeddings of the words in the left and right contexts of $x_i$ within the sentence $s$, including $x_i$ itself. $\overrightarrow{{r}_{e,w}}$ and $\overleftarrow{{r}_{e,w}}$ output their hidden state and we denote with $[\cdot,\cdot]$ the concatenation operation. All the RNNs used in this work are LSTMs.
}

\rev{The output of this computational block is based on the predictions of an MLP classifier that processes (one-by-one) the contextualized embeddings of the words in the input sentence, computed as in eq.~(\ref{eq:CWE}). The MLP is trained using supervised learning, with a tagging scheme similar to \cite{lample2016neural}. In particular, each word needs to be classified as being the \textit{begin, inside, end} word of either an \textit{entity} or a \textit{relation} mention, for a total of 6 classes}. 
As a result, each mention $z$ is composed by the sequence of words where the first word is tagged with the \textit{begin} tag, the last word with the \textit{end} tag, and the other words are predicted as  \textit{inner} \rev{(tags must be all of the same type, either entity or relation)}.

\rev{As remarked in Sections \ref{sec:intro} and \ref{sec:problem}, we exploit sparse supervisions, so that training the MLP-tagger (and, in turn, the embedding networks ${r_{c,w}},{{r}_{e,w}}$) might be difficult. However, we follow the intuition that syntax has a crucial role in text segmentation: noun phrases are mentions to entities, while fragments that start with a verb and end with a preposition (if any) are mentions to relations \cite{fader2011identifying}.\footnote{See the supplemental material for the details.} Hence, we can use these rules to automatically generate artificial supervisions on large collections of text to  pre-train the \textsc{Mention Detector}. }

\subsection{Mention and Context Encoding}
\label{sec:enc}
\rev{Once mentions have been detected, the \textsc{Encoder} module (second block of Figure \ref{fig:flow_chart}) encodes them and their contexts into vectorial representations, following an encoding scheme similar to the one in Section~\ref{sec:seg}. However, after the processing of the previous computational block, the input sentence becomes a sequence of mentions, each composed of one or more words.
}
\rev{In details, at this stage, $s$ is a sequence of $m_s$ mentions $z_i, i=1,\ldots,m_s$. Two different vectorial representations are computed for each mention. First, the \textit{mention embedding} $\Be_i^{m}$ of $z_i$ is obtained as
\begin{equation}
	\label{eq:ME}
	\Be_i^{m} = \overleftrightarrow{r_{c,m}}(\hat{\Bc}_{i,1} \ldots \hat{\Bc}_{i,|z_i|}) \ ,
\end{equation}
being $\hat{\Bc}_{i,1} \ldots \hat{\Bc}_{i,|z_i|}$ the sequence of characters of the mention.}

\rev{Second, the \textit{context embedding} $\hat \Be_i^{m}$ is computed from the other mentions $z_j$, $j\neq i$ in $s$, as
\begin{equation}
\label{eq:CE}
	\hat{\Be}_i^{m}  = [\overrightarrow{{r}_{e,m}}(\Be_{1}^{m} \ldots, \Be_{i-1}^{m}), \overleftarrow{{r}_{e,m}}(\Be_{n}^{m}\ldots \Be_{i+1}^{m})].
\end{equation}
Notice that, differently from eq.~(\ref{eq:CWE}), here we are encoding \textit{only} the context around the considered mention, excluding the mention itself. 
}

In order to learn the parameters of the encoders ($r_{c,m}$ and $r_{e,m}$), we follow the principles at the basis of the Word2Vec-CBOW architecture \cite{cbow}, that is rooted on the idea of encoding the context around a word and decoding the word itself, to generate representations where synonyms or semantically similar words are close to each other in the embedding space.
However, differently from the original CBOW, we use character-level input encoding (Eq. \ref{eq:ME}), so that we also expect morphologically similar inputs to be close in the embedding space. We also introduce another feature that makes our approach different from other related systems (e.g. \cite{charEMB_COMPARISON,marra2018unsup}), that is we keep also the decoding stage at a character-level, thus avoiding the need of a large mention vocabulary.

%
\rev{In particular, once the context around mention $z_i$ is encoded into the context embedding $\hat \Be_i^{m}$ (Eq. \ref{eq:CE}), we exploit an LSTM-based decoder whose initial state is set to $\hat \Be_i^{m}$, and that generates the sequence of characters that compose $z_i$. The encoder and the decoder can be trained by exploiting the cross-entropy loss function over character predictions.}
Once the \textsc{Mention Detector} has been pre-trained accordingly to what suggested in Section \ref{sec:seg}, the \textsc{Encoder} module can be pre-trained as well without any human intervention, by processing large collections of text and learning to decode each mention. 


\subsection{Candidate Generation}
\label{sec:cg} 
\begin{figure}
    \centering
    \hskip -8mm
    \includegraphics[scale=0.3, trim={0 2cm 0 2.5cm}]{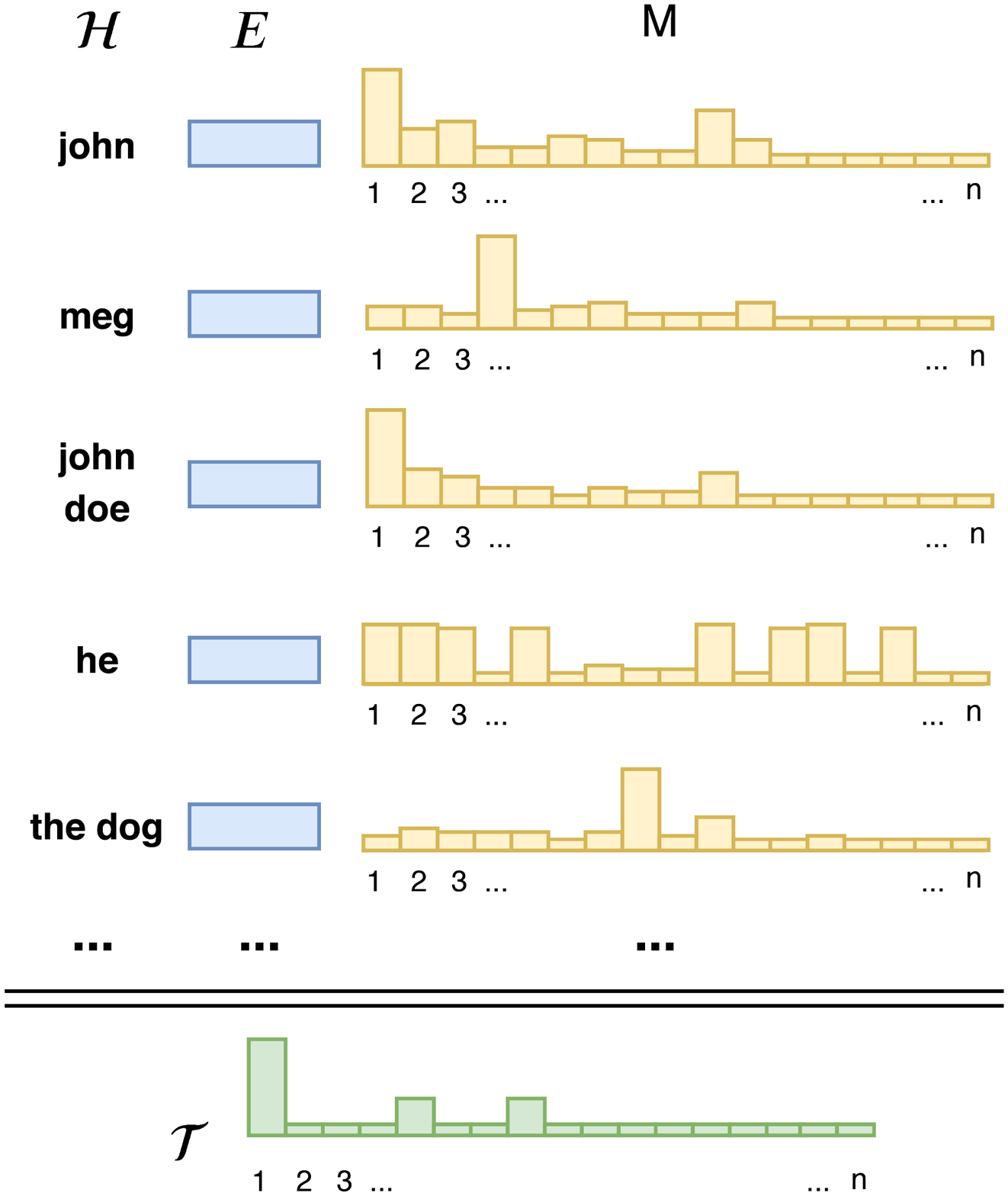}
    \caption{\rev{A graphical representation of the four memory components $\mathcal{H}$, $E$, $M$, and $\mathcal{T}$. $\mathcal{H}$ collects the (lowercase) mentions that are stored in the KB. For each of them, the embedding vector (blue rectangle) is stored in a row of the matrix $E$. The affinity scores of the considered mention, with respect to each of the $n$ instances in the KB, are stored in a row of the matrix $M$. Finally, $\mathcal{T}$ contains the KB instances that have been recently linked by the system (here represented as a histogram of the number of times each KB instance was recently linked).}}    
    \label{fig:memories}
\end{figure}

Given an input mention $z$ from the current sentence and its embedding $\hat \Be^{m}$, \rev{the \textsc{Candidate Generator} (third block of Figure \ref{fig:flow_chart})} implements four \textit{memory components} that are used to generate a list of candidate KB instances compatible with $z$, and, afterwards, to store the information on the disambiguated instance. Before providing further details on the candidate generation process, \rev{we describe the four memory components, as shown in Figure \ref{fig:memories}}. \\

\vskip -4mm$i)$ The memory component $\mathcal{H}$ is an ordered set that collects all the mentions that were processed up to the current sentence; this allows a fast lookup of previously predicted instances for specific mentions. For example, if the mention \textit{John Doe} was previously assigned to instance $k$, when processing the same mention again the system could easily hypothesize that it still belongs to instance $k$.\\
\vskip -4mm$ii)$ The matrix $E$ stores (row-wise) the embeddings of the mentions in $\mathcal{H}$, computed with the encoder of Section \ref{sec:enc}, \rev{Eq. \ref{eq:ME}}. Thanks to a similarity measure in the embedding space, this component allows the system to associate KB instances to never-seen-before mentions which are small variations of previously seen ones, or that refer to semantically similar elements. For example, given the never-seen-before mention \textit{John D.}, the system could easily predict it still belongs to instance $k$, since its char-level embedding is close to the one of \textit{John Doe}, even though the exact lookup in $\mathcal{H}$ failed.\\
\vskip -4mm$iii)$ The set $\mathcal{T}$ keeps track of the last disambiguated instances (with repetitions). This memory naively allows the system to handle co-references. As we will see shortly, the system can learn that some specific inputs (e.g. pronouns, category identifiers, etc.) are often assigned to recently mentioned instances, making valuable temporal hypotheses when it has to disambiguate such inputs.\\
\vskip -4mm$iv)$ The matrix $M$ stores (row-wise) the instance-activation scores of each mention in $\mathcal{H}$. Each row is associated to a mention in $\mathcal{H}$, each column corresponds to a KB instance. The row of $M$ associated to a certain mention $u\in\mathcal{H}$ models how strongly each KB instance is associated to the mention $u$. \rev{The matrix $M$ is learned while reading the text stream, as we will describe in Section \ref{sec:learning}}.

\begin{figure*}
    \centering
    \includegraphics[width=0.8\textwidth,trim={0 0.3cm 0 0}]{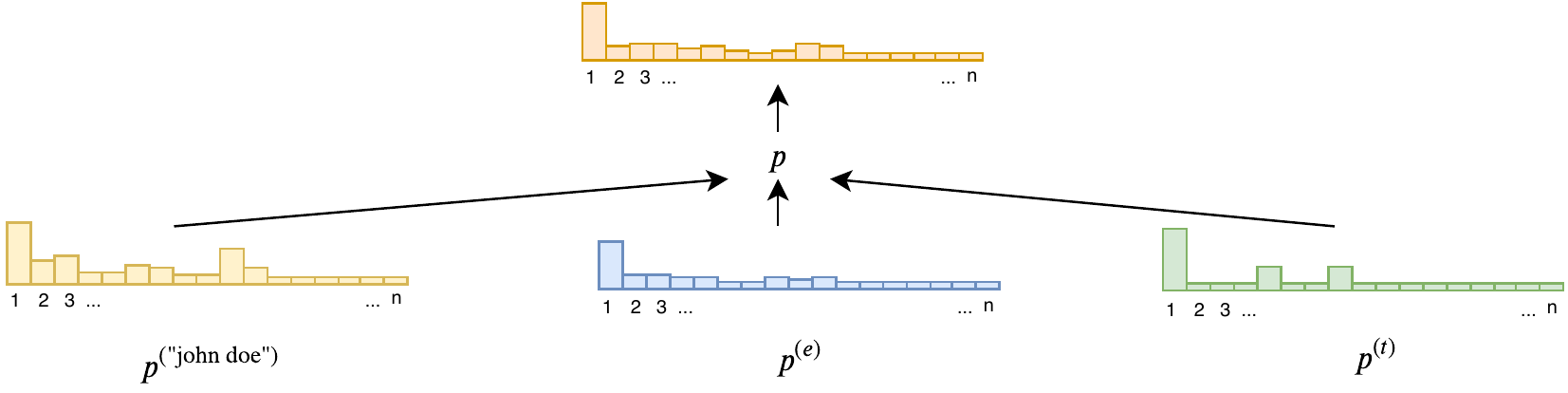}
   \caption{\rev{{A visual representation of the combination of the three hypotheses.} $\textbf{p}^{(z = \text{``john doe''})}$ suggests that the current mention (i.e. ``john doe'') has been often linked to entity instance 1. $\textbf{p}^{(e)}$ indicates that the embedding of the current mention is very similar to the embeddings of mentions usually linked to entity 1. Finally, $\textbf{p}^{(t)}$ indicates that entity 1 has been recently mentioned several times.}}    
    \label{fig:combine_hps}
\end{figure*}

In the following, we will show how the \rev{\textsc{Candidate Generator}} exploits these memory components to generate candidates of KB instances, given an input mention. For the purpose of this description, we suppose that the current memory consists of $n$ instances and $m$ mentions. 
The \textit{candidate generation} routine first outputs three hypotheses, that are represented by three vectors $\Bp^{(z)}$, $\Bp^{(e)}$, $\Bp^{(t)}$, of $n$ scores each. Each score is in $[0,1]$, and it represents how strongly an instance is a candidate for being linked to the mention $z$. For example, $p_i^{(z)}$ (the $i$-th component of $\Bp^{(z)}$) models the probability of the instance $i$ to be a link candidate accordingly to the first hypothesis. Then, the three hypotheses are combined into a final vector $\Bp$. \rev{A visual representation of this combination is shown in Figure~\ref{fig:combine_hps}}.

The first vector, $\Bp^{(z)}$, named \textsc{string-match hypothesis}, contains the activation scores of the $n$ instances given the string $z$. It models the idea that the \textit{surface form} of $z$ is a strong indicator for spotting candidate links. Formally,

\begin{equation}
	\Bp^{(z)} = \sigma \left(M_{h(z,\mathcal{H})} \right) \ ,
\end{equation}
where the function $h(z,\mathcal{H})$ returns the index of $z$ in $\mathcal{H}$, $\sigma(\cdot)$ is a sigmoidal function that operates element-wise on its input (yielding output values in $(0,1)$) and the subscript after the matrix M indicates the matrix row.
 
The second hypothesis, $\Bp^{(e)}$, named \textsc{embedding-match hypothesis}, collects the activations of the instances given the embedding $\rev{\Be^m}$ of $z$. The rationale behind it is that if $\rev{\Be^m}$ is similar to an embedding of a known mention (in the sense of the cosine similarity $\cos(\cdot, \cdot)$), then it is likely to activate the same instances. Due to the way our embeddings are computed (Section \ref{sec:enc}), we expect that two embeddings are ``close'' if they have similar roles in the processed sentences (semantic-similarity, synonymy), and similar morphological properties (due to the character-level input). Formally:

\begin{equation}
\Bp^{(e)} =\left( \left[ \frac{\cos(\rev{\Be^m},E_i)+1}{\sum_{j=1}^{m} \cos(\rev{\Be^m},E_j) + m} \right]_{i=1}^{m} \right)'  \cdot  \sigma(M)
\end{equation}
where the notation $[v_i]_{i=1}^{m}$ indicates the (column) vector $[v_1, \ldots, v_m]$. Notice that $\sigma(M)$ is a matrix of the same size of $M$, and $\Bp^{(e)}$ involves a vector-times-matrix operation that basically computes a weighted sum of the rows of $M$ accordingly to the similarity between $\Be^m$ and the stored embeddings\footnote{In our implementation, we kept only the top-$k$ cosine similarities $(k=5)$, forcing the other ones to $-1$.}.

The third hypothesis, $\Bp^{(t)}$, named \textsc{temporal hypothesis}, implements the idea that recently disambiguated instances are good candidates for co-reference resolution (temporal locality). In other words, if a story is talking about a certain entity, it is likely that the narration will make references to it using some new surface forms. For example, given an entity labeled \textit{Donald Trump}, there could be ambiguous mentions like \textit{Donald}, \textit{Mr. Trump} (other people called this way), or \textit{the president} (that cannot be captured by the other two hypotheses). In these cases, the temporal locality has a crucial role, that is even more evident when using pronouns, that are shared by several instances. Formally, we have
\begin{equation}\label{eq:tp}
\Bp^{(t)} = \frac{ \left[ u(i,\mathcal{T}) \right]_{i=1}^{n}}{\max \left[ u(j,\mathcal{T}) \right]_{j=1}^{n}} \ ,
\end{equation}
where $u(j,\mathcal{T})$ returns the number of occurrences of instance $j$ in $\mathcal{T}$. 

The \rev{\textsc{Candidate Generator}} merges the three hypotheses and produces the \textsc{final hypothesis} $\Bp$. The idea behind this operation is to give more priority to $\Bp^{(z)}$ than to $\Bp^{(e)}$ when the former is strongly activated (since $\Bp^{(z)}$ is about ``exact'' matches in terms of surface forms). The importance of the temporal component $\Bp^{(t)}$ in the merge operation depends on the current mention $z$. For example, if $z$ is a pronoun, the system must  trust $\Bp^{(t)}$ more than the others, while, in case of some unambiguous mention, it must learn that $\Bp^{(t)}$ is not important. We let the system learn the importance of $\Bp^{(t)}$ depending on the mention embedding $\rev{\Be^m}$. Formally, the system computes $\gamma = q(\rev{\Be^m}) \in [0,1]$, where $q(\cdot)$ is a learnable function (whose form will be defined shortly), and the vector of merged hypotheses is
$$
\Bp = \left(1 - \gamma\right) \cdot \left(\Bp^{(z)} + \left(1 - \Bp^{(z)} \right) \Bp^{(e)} \right) + \gamma \cdot \Bp^{(t)} \ .
$$
The vector $\Bp$ contains a set of scores that model how strongly each KB instance is related to the current input mention.

We kept the model as simple as possible by considering only the three hypotheses that were necessary to cover all the ambiguities present in the task at hand. However any number of hypotheses can be attached to the candidate generation module, making our model adaptable to different NLP problems.

\subsection{Disambiguation}
\label{sec:du} 
While the candidate generation routine only focusses on the mention $z$, \rev{the \textsc{Disambiguator} (Figure \ref{fig:flow_chart}, fourth block) is responsible of determining what are the most likely KB instances given the context of $z$. The representation of the context $\hat{\Be}^m$ is computed by the \textsc{Encoder} (Section \ref{sec:enc}) by Eq.~(\ref{eq:CE})}. The \textsc{Disambiguator} is based on the functions \rev{$d_i(\hat{\Be}^m)$, $i=1,\ldots,n$}, also referred to as \textit{disambiguation units}. In details, each $d_i$ is associated to a KB instance, and it is learned while reading the text stream in a supervised or unsupervised way, as we will describe in Section \ref{sec:learning}. \rev{In particular, in the considered problem,} we do not have the use of any discriminative information: when we receive the supervision that the input mention is about a certain KB instance (or when the model decides this in an unsupervised manner), we cannot infer that the context of the considered mention is not compatible with any other KB instance. As a matter of fact, the same context may be shared by several instances, so each $d_i(\cdot)$ must have the capability of learning from positive examples only. For this reason we implement $d_i$ with a locally supported similarity measure,
\begin{equation}
d_i(\rev{\hat{\Be}^m}) = \frac{1}{2} + \frac{1}{2} \max_{j=1,\ldots,\kappa} \cos (\rev{\hat{\Be}^m}, \Bw_{ij}) \in [0,1] \ ,
\label{eq:d}
\end{equation} 
that models the distribution of the contexts for instance $i$ by means of $\kappa$ centroids $\{\Bw_{ij}\}$. \rev{As we will describe Section \ref{sec:learning}, these centroids are developed in an online manner}, and, in the unsupervised case, we end up in an instance of online spherical K-Means.
Also the previously mentioned function $q(\cdot)$ needs to be locally supported (for the same reasons), so we implemented it following Eq. (\ref{eq:d}) as well.

We combine the activations of candidates (i.e., the vectors $\Bp$), and the disambiguation-unit outputs, $\Bd = [d_i(\hat{\Be})]_{i=1}^{n}$, to get the output $\Bo$ of the system,
\begin{equation}
\Bo = \delta(\Bp > \tau_r) \cdot \left( \eta \cdot \Bp + (1-\eta) \cdot \Bd \right) \ ,
\label{eq:out}
\end{equation} 
where $\delta$ returns a binary vector with 1's in those positions for which the (element-wise-evaluated) condition in bracket is true. The scalar $\tau_r > 0$ is a \textit{reject} threshold, and $\eta \in [0,1]$ is a tunable parameter that controls the role of the hypothesis $\Bp$ in the decision process\footnote{For simplicity, we set $\eta=0.5$.}. The reject threshold allows us to avoid computing the $d_i(\cdot)$ associated to very-low-probability candidate instances.

%% file: learning.tex
The system reads data from a text stream and it optimizes the model parameters in an online manner. Before going into further details, we recall that the learnable parameters of the proposed model are the matrix $M$ in the memory component, the vectors $\Bw_{ij}$ of the disambiguation units and of the temporal relevance function $q$ (\rev{i.e., the parameters of the \textsc{Candidate Generator} and of the  \textsc{Disambiguator}}), and the weights of the LSTMs in the \rev{\textsc{Mention Detector} and in the \textsc{Encoder}}.

The system starts with an empty KB (so $M$ is not allocated yet), and with randomly initialized model parameters.
As already introduced in Sections \ref{sec:seg} and \ref{sec:enc}, we can pre-train those modules that constitute the preliminary stages in the system pipeline of \rev{Figure \ref{fig:flow_chart}, i.e., the \textsc{Mention Detector} first, and then the \textsc{Encoder}} (in both cases, without human intervention). In other words, the system will start reading data from the text stream and it will progressively acquire the skill of detecting mentions to entities and relations, and the skill of encoding such mentions and their context. When a stationary condition is reached in the detector and encoders, the system can start to develop also the KB-based disambiguator, and to eventually refine and improve the pre-trained modules.

\begin{algorithm}
\footnotesize	
\ForEach{mention in the current sentence} {   
		 \If{supervision not provided}{
			\If{recognized some instances}{
				reinforce the associated disambiguation units
				}
			\ElseIf{uncertain disambiguation}{
					no actions
				}
			\ElseIf{no recognized instances}{
					create new instance\\
					reinforce the new disambiguation unit until $>\tau_a$
				}	
			
		 }\Else{
		 	\If{already known supervision}{
				reinforce the associated disambiguation unit until $>\tau_a$\\
				penalize the other disambiguation units until $<\tau_a$
			}\Else{
			 	disambiguate instance $y$\\
				\If{$y$ was associated to another supervision}{
					goto line 10
				}\Else{
					associate supervision to the $y$-th disambiguation unit\\
					goto line 16
				}
			}
			
		 }
%
%

	}
	\caption{Learning Dynamics}\label{alg:ld}
\end{algorithm}

While processing the text stream, accordingly to Eq. (\ref{ooo}), each detected mention $z$ is associated to a disambiguated KB instance $y$. 
Before starting the disambiguation, the system verifies if $z$ is already in $\mathcal{H}$. If it is not the case, then $z$ is included in $\mathcal{H}$, its embedding \rev{$\Be^m$} is appended to $E$, and a new row is added to $M$ (with values such that $\sigma(M_{h(z,\mathcal{H})}) \approx 0$).
The learning stage consists in an online process to optimize the model parameters accordingly to either \textsc{self-learning} or a \textsc{supervision} about the target instance $\overline{y}$. A sketch of the whole learning stage is shown in Algorithm \ref{alg:ld}.

\textbf{Self-Learning}. When no supervision is provided, learning dynamics changes in function on the confidence that the system yields in formulating hypotheses ($\Bp$) and in disambiguating the mention $z$ ($\Bo$). We distinguish among three cases:
\begin{itemize}
\item[\textit{i.}] $\max \Bo \geq \tau_a:$ \textsc{\footnotesize recognized some instances}
\item[\textit{ii.}] $ \max \Bp > \tau_r \land \max \Bo < \tau_a:$ \textsc{\footnotesize uncertainty}
\item[\textit{iii.}] $\max \Bp \leq \tau_r:$ \textsc{\footnotesize unknown instance} \ ,
\end{itemize} 
where $\tau_r$ is the aforementioned \textit{reject} threshold, and $\tau_a \in (\tau_r, 1)$ is an \textit{accept} threshold.\footnote{We set $\tau_r=0.1$ and $\tau_a=0.9$.}
In case \textit{i.} the response of the system has been rather strong in indicating at least one instance, therefore the prediction of the model is considered reliable enough, so such decision must be reinforced for all those disambiguation units that generated outputs in $\Bo$ above the \textit{accept} threshold $\tau_a$. 
This is done in a self-learning fashion \cite{nigam2000analyzing}, by means of a single online gradient-based update, with the aim of minimizing the quadratic loss that measures the distance between the selected disambiguation unit outputs (indexed by $j$) and $1$, that is $\sum_j(d_{j}-1)^2$ (i.e., we reinforce the selected outputs).
In case \textit{ii.} the system activates some candidates but it is uncertain in the disambiguation, so no further actions are taken. 
Case \textit{iii.} is triggered when $\Bp$ is composed of only low-confidence candidate activations. This situation happens when the candidate generation module does not find a known instance that is compatible with the current mention, that is likely to indicate the occurrence of a new entity/relation. Therefore, the system creates a new instance in the KB and reinforces its disambiguation unit until its response is above $\tau_a$, to develop the new instance model (i.e., multiple gradient-based updates). 


\textbf{Supervision}. When a supervision $\overline{y} \in \mathcal{\overline{Y}}$ is provided for the mention $z$, we want the system to immediately learn from it. 
The system keeps track of the mapping between the set of user-provided supervisions $\mathcal{\overline{Y}}$ and the set of instances in the memory components (the user is not aware of how the system handles instances). When $\overline{y}$ is a known supervision, the index of the corresponding instance is found, and the output of the disambiguation unit associated to this instance is reinforced until it is greater than the accept threshold $\tau_a$. We also push the output of the other disambiguation units towards $0$, by minimizing the quadratic loss, $\sum_j(d_{j})^2$. This implements a penalization process, that involves multiple gradient-based updates until all the involved outputs fall below $\tau_a$\footnote{Notice that, whenever the system needs to reinforce the $j$-th output $o_j$, and it also holds that $p_j \leq \tau_r$, then, due to $\delta(\cdot)$ in Eq. (\ref{eq:out}), we get no gradient, so we first increase $p_j$ until it is above $\tau_r$.}.
When $\overline{y}$ is a never-seen-before supervision for the system, then it is associated with the disambiguated instance $y$. On the other hand, if $y$ was previously associated to another supervision symbol in $\mathcal{\overline{Y}}$ different from $\overline{y}$, then we have a collision in the mapping and we solve it by creating a new instance and by associating it to $\overline{y}$. Then, we follow the same steps as in case \textit{iii} above\footnote{Supervisions may not only be related to the instance-label of the detected mentions, but they can also be associated with the detection of the mention. For example, the user could label a mention that does not correspond with the detected ones. This supervision signal is propagated to the mention detector, that can be refined and improved. In turn, the mention and context encoders could be refined as well. The investigation of these refinement procedures goes beyond the scope of this paper.}.

\medskip

%% file: related.tex

Mining over \textit{text streams} has been studied in a number of works \cite{banerjee2007topic,krawczyk2017ensemble,aggarwal2012mining}, with several different purposes, that, however, are different from what we consider in this paper. 
Our approach to the learning problem is based on simple sentences that have the same structure of the ones used in many tasks of the \textit{bAbI project} by Facebook \cite{sukhbaatar2015end,pmlr-v48-kumar16}. However, none of such tasks is conceived for online learning or for entity/relation extraction and disambiguation. 
Interesting ideas on entity-oriented sub-symbolic memory components have been recently proposed by \cite{henaff2016tracking,dynamic2017}, and extended to the case of relations by \cite{bansal2017relnet}: their formulation is developed to comply with the aforementioned bAbI asks.
The idea of considering small text passages could resemble the task of \textit{Machine Comprehension}, where, however, such passages are read with the purpose of answering a question \cite{richardson2013mctest,rajpurkar2016squad,kobayashi2016dynamic}. 
Concerning the input representation, we took inspiration from those works that exploit character-level embeddings to build models that also take into account morphological information \cite{kim2016character,charEMB_COMPARISON}.

Our approach disambiguates mentions using their contexts, so it shares several aspects with \textit{Word Sense Disambiguation} (WSD) and \textit{Entity Linking} (EL), that, differently from our case, assume to work with a given KB.
In WSD \cite{ims,raganato2017word} the set of target words is known and the senses of each word are taken from a given dictionary. 
EL \cite{shen2015entity} is similar to WSD \cite{moro2014entity,MoroNavigli:2015}, but it is about linking ``potentially partial'' entity mentions to a target KB, that has an encyclopaedic nature \cite{hachey2013evaluating,MoroNavigli:2015}. The EL problem is presented in several variants and focussing on different types of data \cite{ling2015design,guo2013link,pan2015unsupervised,oneforall2016,pappu2017lightweight,lin2017list}, and it has been the subject of task-oriented evaluation procedures and benchmarks \cite{ma2017cmu,van2016evaluating}. A few EL systems work in an unsupervised way \cite{han2011generative,pan2015unsupervised}, but the KB is still given. 
\textit{Named Entity Recognition} (NER) focusses on discovering mentions to entities, and it is also a basic module of several EL systems \cite{luo2015joint}. However NER is about proper nouns, as frequently is EL \cite{ling2015design}, while here we also consider common nouns. Moreover, NER systems output the entity type (person, location, etc.) without producing any instance-level information \cite{lample2016neural,chiu2016named}. \textit{Relation Extraction} (RE) has been recently approached with end-to-end and advanced embedding-based models \cite{miwaend,obamuyidecontextual}. The entities involved in the target relation are usually known, and a pre-defined ontology is given (distant supervisions are also used, as in \cite{mintz2009distant}). There are a number of discussions to better state the RE problem and build accurate gold labels \cite{martin2016callRECAOS}.


Finally, learning the KB component is the subject of those tasks of automatic \textit{KB Construction} \cite{niu2012deepdive} and \textit{KB Population} \cite{dredze2010entityKB_POPULATION,ji2011knowledge}, that, differently from the case of this paper, either make some application-specific assumptions to implement the KB, or exploit a given ontology schema, also combining unsupervised and supervised learning with ensembles and stacking techniques \cite{rajani2016combining}. 

%% file: exp.tex

\subsection{\rev{Datasets}}
\label{subsec:datasets}

\begin{figure}[!ht]
	\centering
	\includegraphics[width=0.5\textwidth,clip]{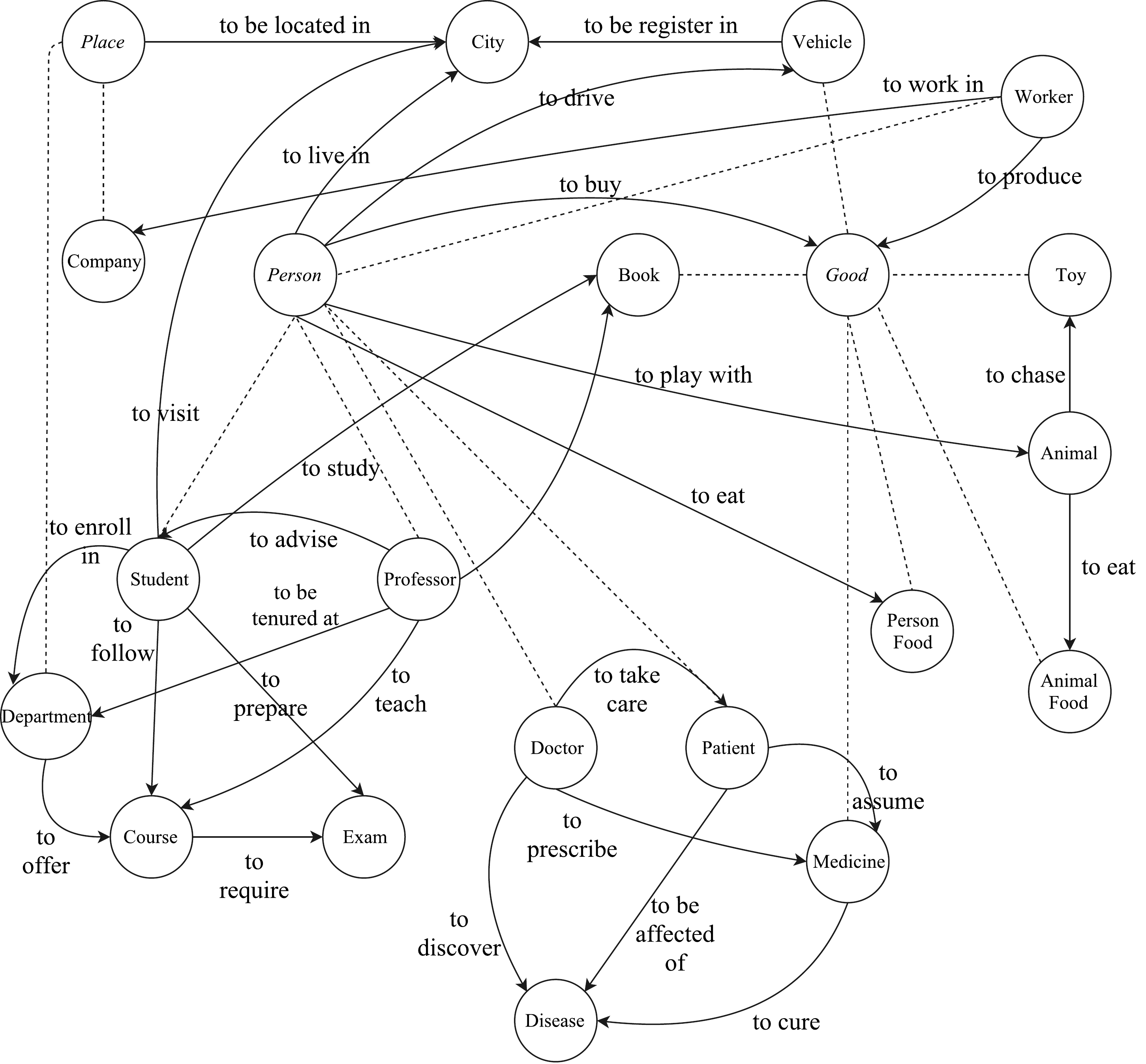} 
	\caption{Ontology Graph of entity (nodes) and relation (edges) types in the Simple Story Dataset. Dashed lines between node pairs indicate a hierarchy between the two types.}
	\label{fig:ontology_graph}
\end{figure}

\noindent \textbf{Simple Story Dataset.} A detailed experimentation was carried out in a new dataset that we created and made publicly available at \url{ http://sailab.diism.unisi.it/stream-of-stories/}.
We remark that the problem we face shares some aspects with existing benchmarks, but none of them is really focussing on what we introduced in Section \ref{sec:problem}.
The data we created is composed of a stream of 10,000 sentences, organized into 564 stories (similarly to what reported in Figure \ref{fig:ex}). Each story is composed of a list of not-repeated facts, involving 130 entity and 27 relation instances that belong to a pre-designed ontology (\textit{not provided to the system}) shown in Figure \ref{fig:ontology_graph}. Facts in a story mostly talk about a certain entity, that we refer to as ``main entity'', and that can also appear in other stories. Entities and relations are mentioned with different surface forms (synonyms, sub-portions of names, etc.). 


In particular, a sentence is constituted by a triple of mentions, involving two entities and a relation.
We automatically generated the data, after having defined the aforementioned ontology with $21$ and $28$ entity and relation types, respectively (Figure \ref{fig:ontology_graph}). Different kinds of noise are introduced, from character-level perturbations (simulating typos) to non-main-entity related story facts (to make descriptions slightly depart from the main subject). The resulting dataset consists of $2176$ unique single word tokens, and dictionaries of $1526$ and $288$ mentions to entities and relations, respectively, that include different variations (typos, determiners, etc.) of $354$ base entity mentions and of $176$ base relation mentions. There are $7975$ co-references (including pronouns). Finally, $6830$ mention occurrences are ambiguous (i.e., refer to multiple instances).

\begin{figure*}
	\includegraphics[width=0.246\textwidth,trim={1.0cm 0.4cm 1.0cm 0cm},clip]{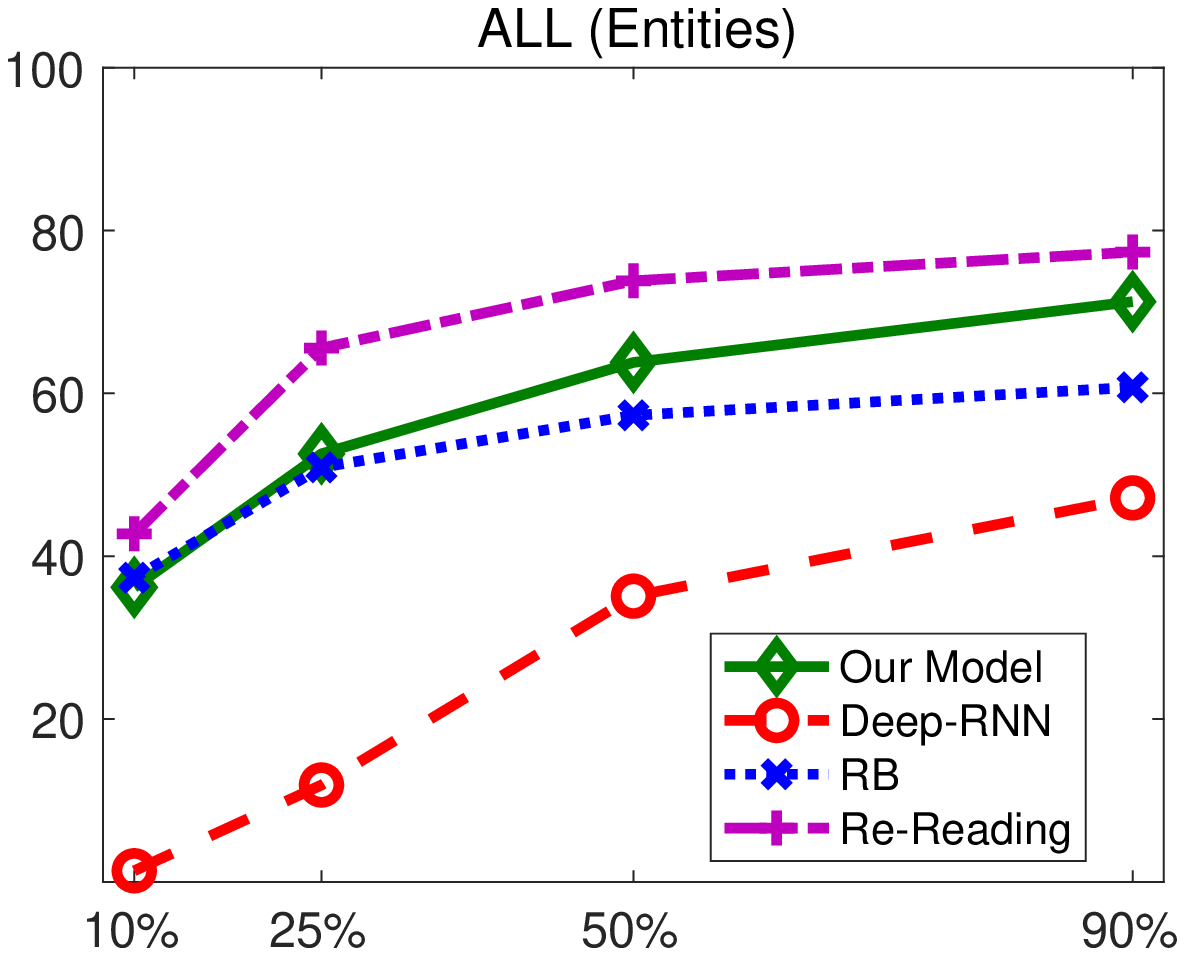}
	\includegraphics[width=0.246\textwidth,trim={1.0cm 0.4cm 1.0cm 0cm},clip]{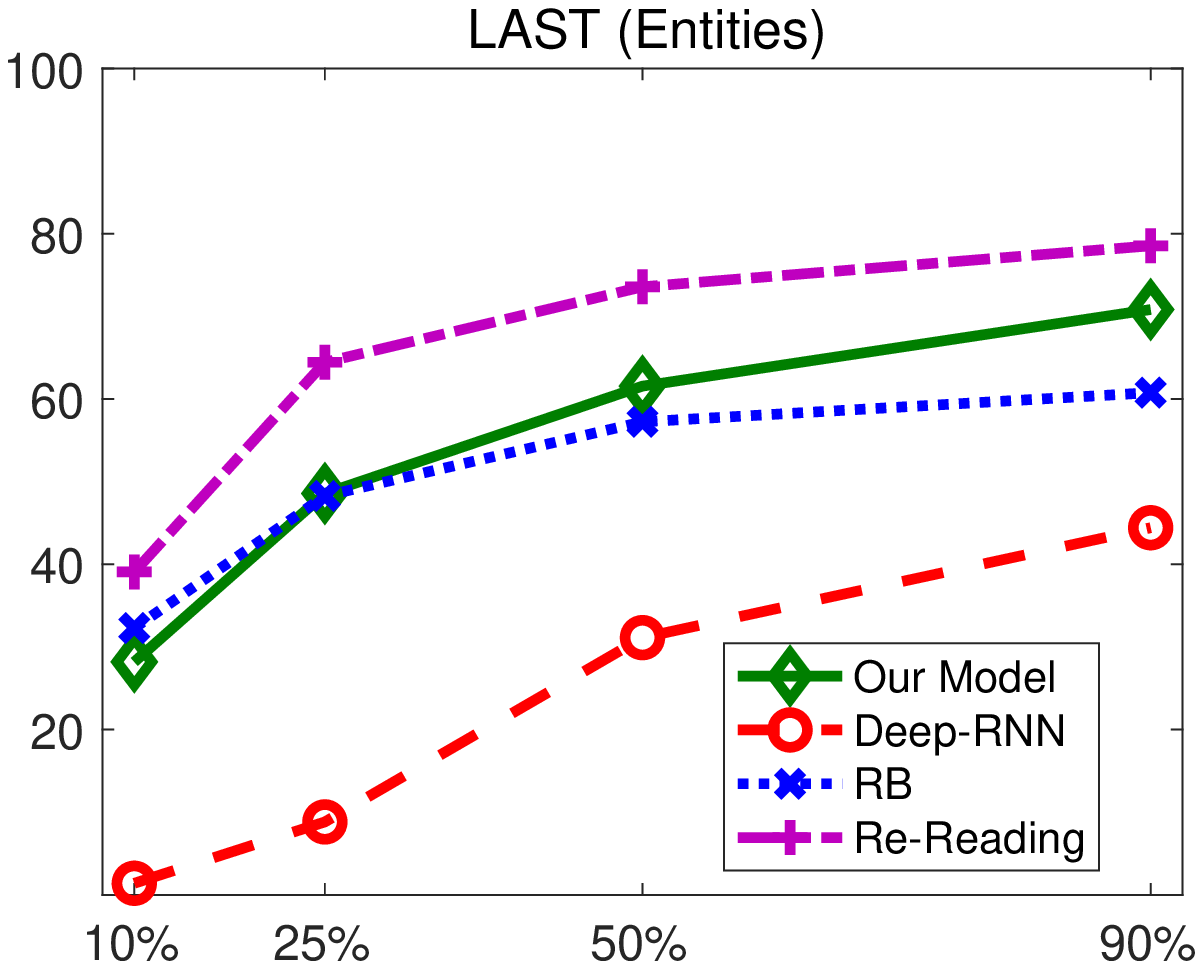} 
	\includegraphics[width=0.246\textwidth,trim={1.0cm 0.4cm 1.0cm 0cm},clip]{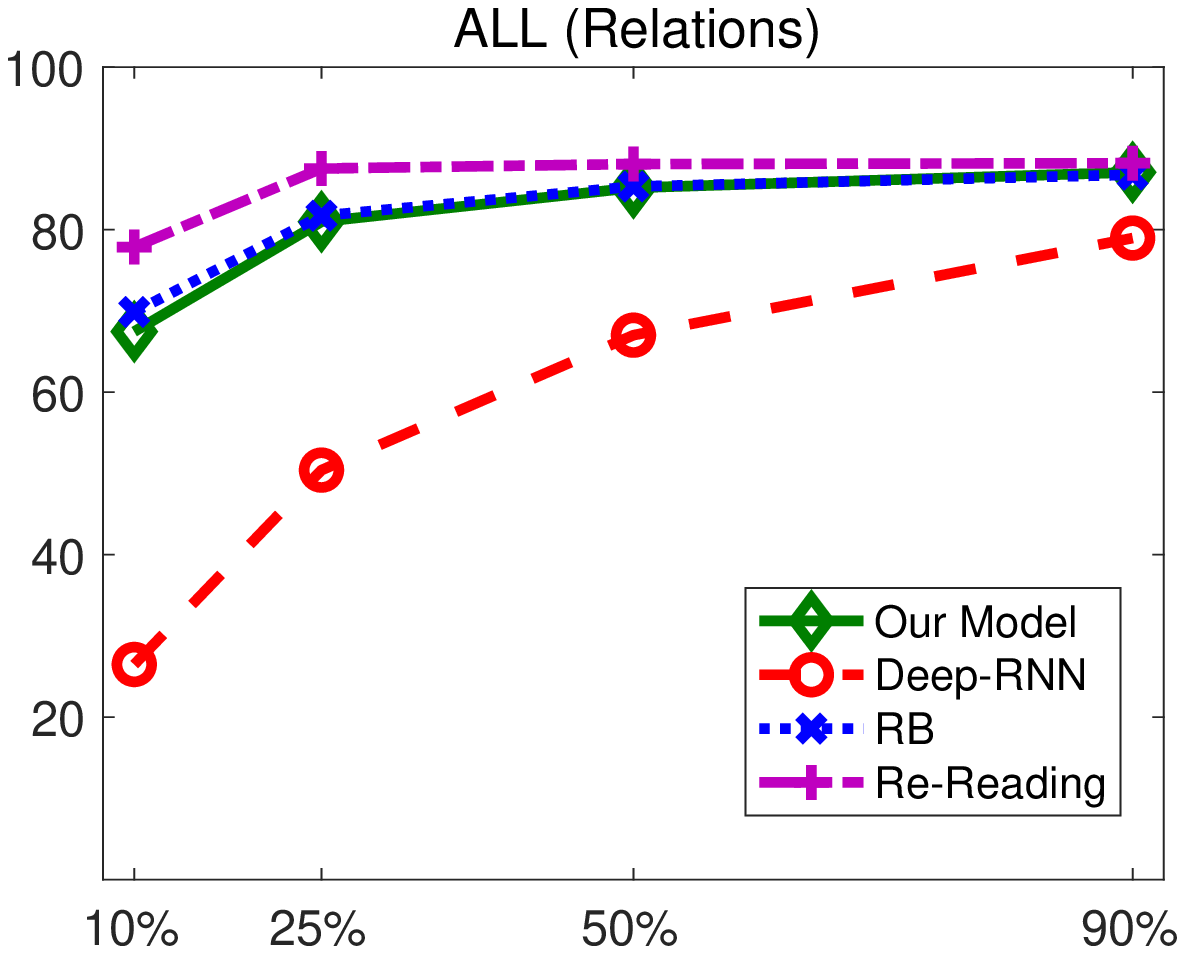}
	\includegraphics[width=0.246\textwidth,trim={1.0cm 0.4cm 1.0cm 0cm},clip]{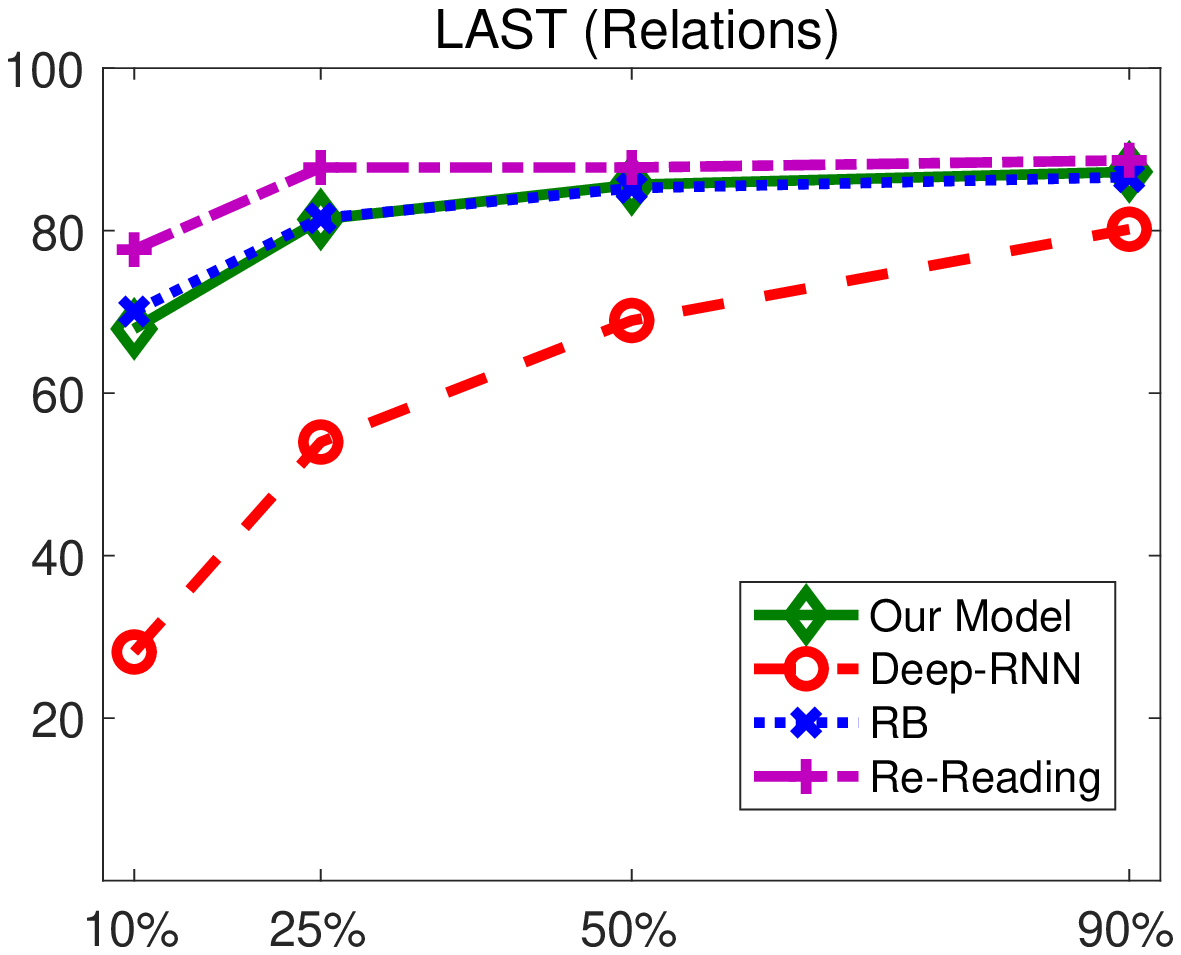}
	\caption{Accuracy ($\%$) for different amounts of supervision in the Simple Story Dataset, in the case of entities (two leftmost graphs), and relations. We include two competitors (\textit{Deep-RNN}, \textit{RB}), our system, and our system reading the stream again.}
	\label{fig:all_last_E_R}
\end{figure*}

%
%
%
%
%

\begin{table}
\centering
\caption{\rev{Accuracy ($\%$) for different amounts of supervision in the Wikifacts Dataset.}}
\rev{
\begin{tabular}{llllll}
\toprule
& Model & $10\%$ & $25\%$ & $50\%$ & $90\%$ \\
\midrule
 \multirow{4}{*}{\textbf{All}} & RB         & $16.84$ & $40.44$ & $48.28$ & $49.55$ \\
& Deep-RNN       & $0.6$  &   $3.01$   &   $12.34$   &   $21.78$   \\
& Our Model  &  $39.25$    &  $\textbf{54.57}$    &  $\textbf{69.64}$    &     $\textbf{75.45}$  \\
& Re-Reading &  $\textbf{44.75}$    &  $\textbf{54.66}$    &  $66.88$    & $70.55$ \\ 
\midrule
\multirow{4}{*}{\textbf{Last}} & RB     & $17.28$ & $40.87$ & $48.04$ & $49.37$ \\
& Deep-RNN       & $0.6$  &   $3.25$   &   $12.11$   &   $21.37$   \\
& Our Model  &  $37.44$    & $52.93$     &  $\textbf{67.45}$    &   $\textbf{75.37}$   \\
& Re-Reading &   $\textbf{43.41}$   &  $\textbf{53.38}$    &   $65.13$   &  $70.39$ \\ 
\bottomrule
\end{tabular}}
\label{exp:wikifacts}
\end{table}

\begin{figure}
\centering
	\includegraphics[width=0.241\textwidth,trim={1.0cm 0.4cm 1.0cm 0cm},clip]{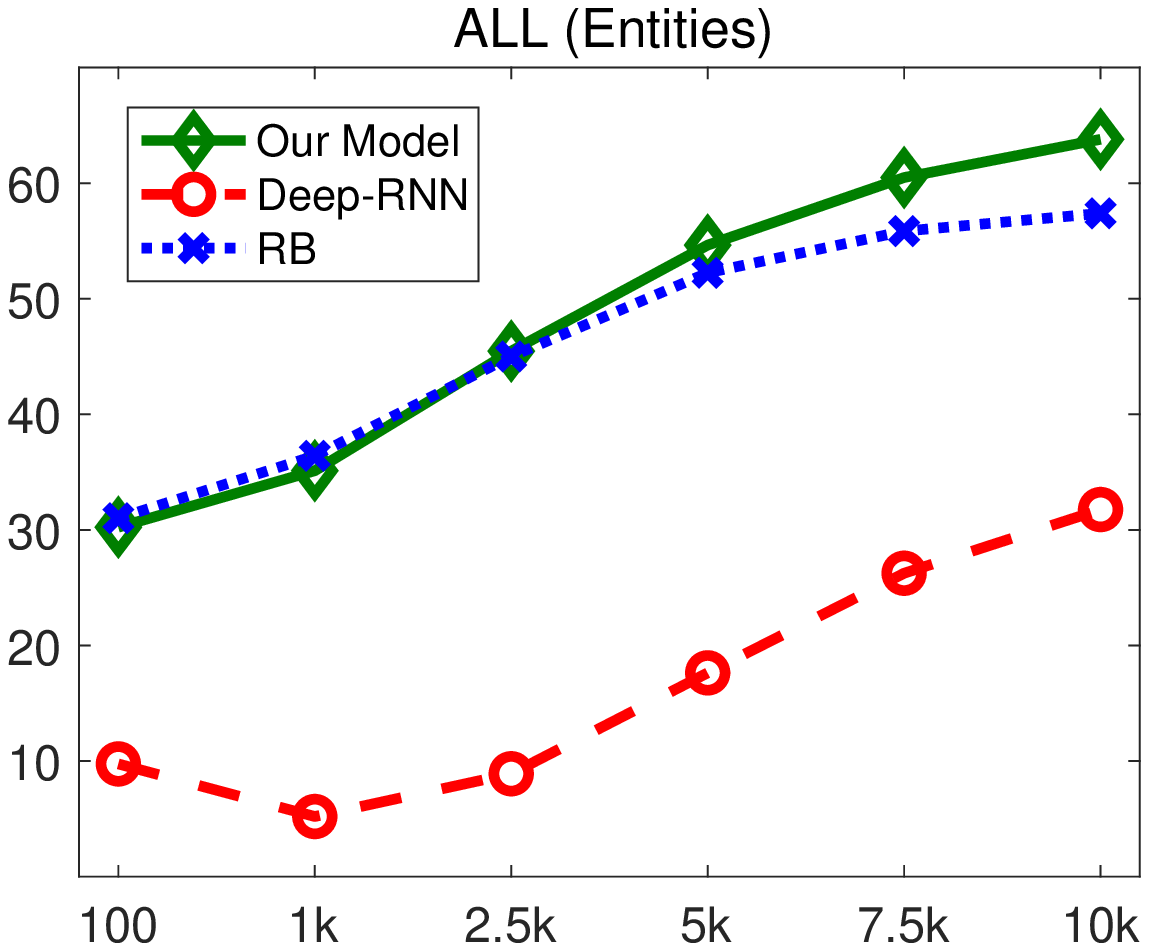}
	\includegraphics[width=0.241\textwidth,trim={1.0cm 0.4cm 1.0cm 0cm},clip]{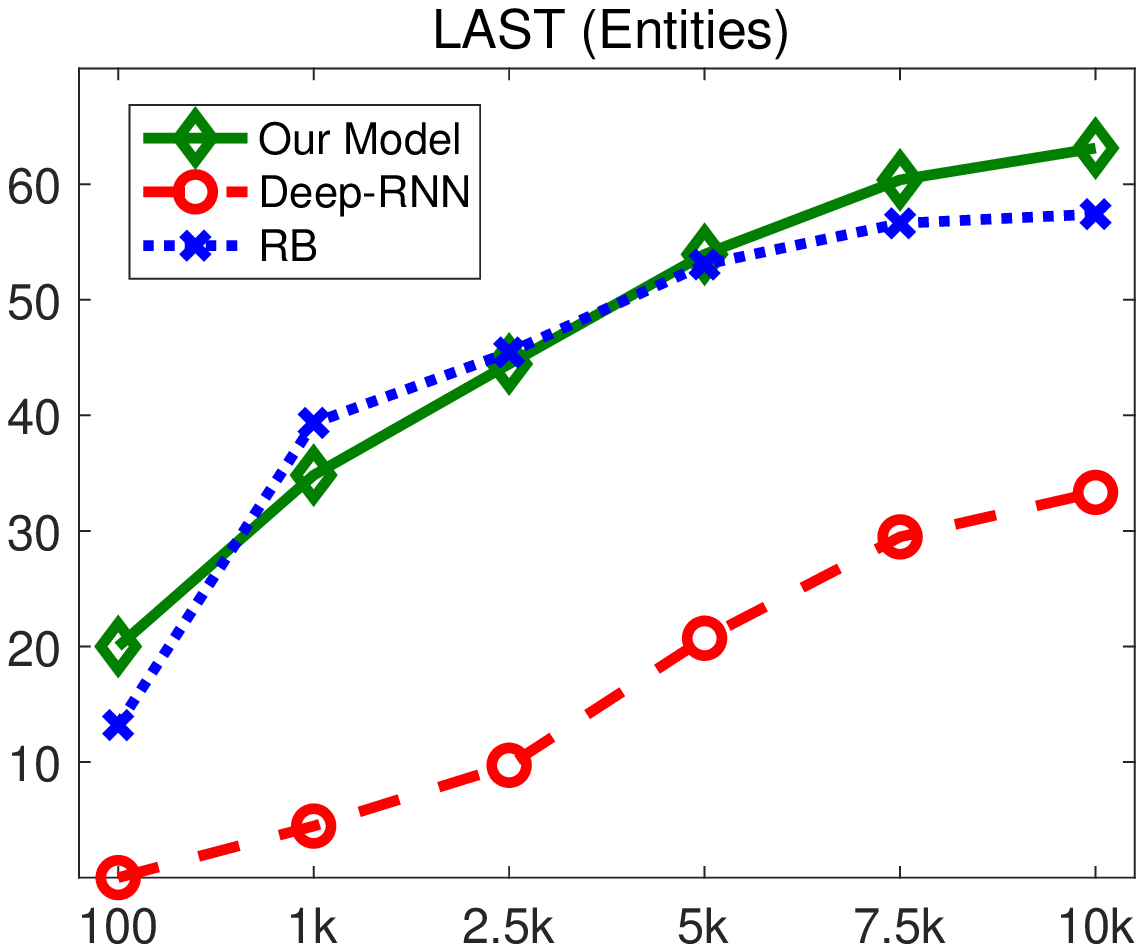} 
	\caption{Accuracy ($\%$) at different time instants for the Simple Story Dataset.}
	\label{fig:time_exps_all}
\end{figure}

\noindent \textbf{WikiFacts.} \rev{WikiFacts is a dataset proposed in \cite{ahn2016neural} where Wikipedia pages are loosely aligned with Freebase triples. It is composed of a collection of summaries, each of them being the textual description of a certain entity, belonging to the domain of movies. The textual span where an entity is mentioned on Wikipedia summaries is annotated with its identifier on Freebase, and the Freebases fact to which it participates. Relations are not explicitly segmented in the text.
Overall the dataset contains about 560k entities extracted from 10k pages. 
WikiFacts can be adapted to the problem setting we consider in this paper (Section \ref{sec:problem}). In particular, we focussed on a subportion of data containing 10k Freebase facts (1112 pages), and we used each summary as a story, keeping only stories longer than three sentences. We considered all the sentences that include at least two entities, and we artificially marked as relation the text span between two consecutive entities. Since relations are not annotated with any Freebase identifiers, we only measure the accuracy on linking mentions to entity instances. A remarkable difference between the previous dataset and WikiFact is that, in the latter, entities are less often repeated among the stories. There are 4431 entities in 10K facts, with respect to the 130 entities in the previous data collection.}


\subsection{\rev{Learning Settings}}
\label{subsec:learning_setting}
We split each story into two parts: a supervised and an unsupervised one.  The supervised portion covers the first sentences of the story, and, in different experimental settings, it consists of $10\%$, $25\%$, $50\%$ and $90\%$ of the story sentences. 
The system reads the data, receives supervisions, and it makes predictions on the unsupervised sentences, accordingly to the stream order. The accuracy on each prediction is measured at the same time when the prediction is made. The results are reported considering all the unsupervised sentences of a story (ALL - this set of sentences is different in function of the supervised part size), and only the last sentence of the story  (LAST - this set is the same for all the supervised part sizes). They are averaged over each story first, and, then, over the whole set of stories. We use the same criterion of the cluster purity measure \cite{manning2008introduction} to map unsupervised outputs to the ground truths, where, in case of conflicting assignments, we only keep the mappings that are determined using the largest statistics. The map used to convert predictions is computed with the statistics accumulated up to the time when the prediction is made.

Our model\footnote{The details of the architecture we selected are reported in the supplementary material.} was bootstrapped accordingly to the scheme of Section \ref{sec:learning}. In particular, before streaming the stories, we generated a stream of text composed of simple language sentences, taken from  Simple English Wikipedia and the Children's Book Test (CBT) data\footnote{\url{https://simple.wikipedia.org}, \url{https://research.fb.com/downloads/babi/}}, getting overall a total of $1.5$ million sentences. We automatically generated supervisions for the mention detector, accordingly to the procedure described in Section \ref{sec:seg}, and we processed the stream. Afterwards, we stopped updating the mention detector and we streamed the same data again to allow the system to develop the mention and context encoders (Section \ref{sec:enc}). In both the cases of the detector and of the encoders, we randomly injected a char-level noise (typos) to make the models more robust to these kind of perturbations. Finally, we stopped updating the encoder and we started to stream the stories. 

\subsection{\rev{Competitors}}
\rev{We compared our model with two competitors. The first one, \textit{Deep-RNN}, is a deep neural architecture, where bottom layers completely coincide with our \textsc{Encoder} module, thus they are two stacked bidirectional RNNs operating on characters and mentions levels, respectively (they are pre-trained as in our model). In order to classify mentions, the  \textit{Deep-RNN} architecture uses an MLP (1 hidden layer with $600$ units and $\mathrm{tanh}$ activation; $\mathrm{softmax}$ activation in the output layer) on top of the concatenated representation $[\Be^{m},\hat{\Be^{m}}]$ (Eq. \ref{eq:ME}, Eq. \ref{eq:CE}). 
This network knows in advance the number of instances of the datasets, that is the size of the output layer of the MLP, and it does not incur in errors related to the self-discovering of new instances. We followed a classic online supervised learning scheme, where a single gradient-based update is performed for each processed sentence, otherwise we experimented that the network simply overfits the last supervisions and forgets about the rest.}
We also considered a very informed \textit{rule-based model}, \textit{RB}, that buffers statistics on the supervisions received up to time $t$. Given an input mention, \textit{RB} predicts the most-common supervision for it. When never-supervised-before mentions are encountered, \textit{RB} predicts the most-frequent supervision of the story, that is likely to be the main entity of the story itself. This information is very precious (our model has no access to it), since several co-references refer to the main entity. We tested a large number of different rule-based classifiers, and \textit{RB} was the one leading to the best results.
Finally, we report another variant of our model, i.e. the case in which the system reads the whole stream another time (``\textit{Re-Reading}'' without providing supervisions again), that is the setting in which the self-learning skills are emphasized.

\subsection{\rev{Results}}\label{subsec:results}
Figure~\ref{fig:all_last_E_R} shows the accuracy of discovering and disambiguating mentions to entity and relation instances, in the Simple Story Dataset, while \rev{Table \ref{exp:wikifacts} shows the accuracy in the case of the Wikifacts dataset (entities).}
Our system outperforms the competitors in all our experiments. Differently from \textit{Deep-RNN}, our model exploits its capability of building local models of the instances, while \textit{Deep-RNN} is not able to capture this locality, either not-learning enough or overfitting supervisions. Moreover, \textit{Deep-RNN} has difficulties in storing information on the whole story, as shown in the LAST case.

\rev{In the case of entities with few supervisions, \textit{RB} shows an accuracy that is similar to the one of our system on the Simple Story Dataset. Differently, the proposed model significantly outperforms RB on WikiFacts, showing interesting generalization capabilities on real world data.}
In the case of relations, \textit{RB} is not different from our system, mostly due to the fact that relations are not so ambiguous in the Simple Story Dataset, also confirmed by the improved results of the \textit{Deep-RNN}. In most of the cases, the classification capabilities of our model  improve when it is allowed to read the stream a second time (re-reading), comparing favourably with all the other approaches. This property is more evident in the few-supervision settings. These results show that, despite the dynamic nature of the online problem we face, the proposed model has strong memorization skills, and the capability of improving its confidence by self-learning.

In the Simple Story Dataset, we also investigate the behaviour of the model at different time instants, using $50\%$ of the supervisions (entities). Basically, we paused the system at a certain point of the stream, measured the accuracy, and activated the system again. This process is repeated until the end of the stream is reached. Results are reported in Figure~\ref{fig:time_exps_all}. Our model reaches better results than \textit{RB} after having read roughly the $20-25\%$ of the input stream. We analyzed this result, and it turned out that this is mostly due to the fact that our system takes some time to learn how to handle the temporal information  ($\Bp^{(t)}$) needed to resolve co-references.
As a matter of fact, co-reference resolution clearly depends on the number of supervisions, as reported in Table~\ref{exp:coref}. 

\begin{table}[!ht]
	\centering
	\caption{Accuracy ($\%$) in the case of pronouns.}
	\begin{tabular}{lllll}
		\toprule
		\textbf{Supervision} & $10\%$  &$25\%$ & $50\%$ & $90\%$ \\
		\midrule
		\textsc{All} 	& $12.74$	&$36.04$	&$43.61$	&$53.55$\\
		\textsc{Last}          &$8.57$	&$42.85$	&$44.76$&	$54.28$ \\
		\bottomrule             
	\end{tabular}
	\label{exp:coref}
\end{table}

We also evaluated the values of $\gamma$, that weighs the importance of the temporal locality in the disambiguation process. In the case of pronouns the average value of $\gamma$ is 0.33, while in the case of the other mentions (that might include some other co-references different from pronouns) is 0.21, thus showing that the system learns to give more importance to the temporal component when dealing with pronouns than other mentions. This is confirmed by the small instance-activation scores in the case of mentions that are pronouns (the average of $\max \Bp^{(z)}$ is 0.005, remarking their inherent ambiguity), with respect to the ones of the other mentions (average of $\max \Bp^{(z)}$ is 0.5).


\subsection{\rev{Ablation Study}}
\rev{We compare different variants of the model in order to emphasize the role of each component}, and report the results in Table~\ref{exp:WE} - Simple Story Dataset. A first comparison regards the benefits of using a character-level encoding for mentions with respect to classical word-level embedding approaches. To this end, we built a vocabulary of words, including all the correct-spelled words of our dataset (and an out-of-vocabulary token), and we limited the character-based encoder to such words, thus simulating a word-level encoder. Our model is able to encode in a meaningful way those words that contain typos, and to exploit them in context encoding, while the word-level encoder faces several out-of-vocabulary words, that also creates ambiguity while comparing contexts.

Another variant of our model discards the temporal hypothesis $\Bp^{(t)}$ when disambiguating entities, and considers the hypotheses $\Bp^{(z)}$ and $\Bp^{(e)}$ only. Table~\ref{exp:WE} shows that the temporal locality has an important role, and disabling it degrades the performances. This is not only due to its positive effects in co-reference resolution, but also when disambiguating mentions to the main entity of the story. Finally, thanks to $\Bp^{(t)}$, the system learns to develop the tendency of associating new mentions to already existing instances, instead of creating new ones, that is an inherent feature of each story (in the worst case it creates only $183$ entity instances).

\begin{table}
	\centering
	\caption{Accuracy ($\%$) of our Full Model, of a system based on Word-Level (WL) Encoding, and of a system not-provided with information on the recently disambiguated instances.}
	\begin{tabular}{lllll}
		\toprule
		\hskip -2mm& \multicolumn{2}{c}{\textbf{Entities}}                         & \multicolumn{2}{c}{\textbf{Relations}}                         \\
		\hskip -2mm & \multicolumn{1}{c}{\textsc{All}} & \multicolumn{1}{c}{\textsc{Last}} & \multicolumn{1}{c}{\textsc{All}} & \multicolumn{1}{c}{\textsc{Last}} \\ \midrule
		
		WL Enc.      \hskip -2mm    & $47.39     $            & $46.18    $           & $84.33   $             & $85.10 $  \\
		No Recent \hskip -2mm & $55.56 $ &$ 54.95$ & $-$ & $- $\\
		Full Model \hskip -2mm & $ \textbf{63.79}   $              & $ \textbf{63.12}$                 & $\textbf{85.41}  $                &$ \textbf{85.81}   $              \\
		\bottomrule                  
	\end{tabular}
	\label{exp:WE}
\end{table}

\minrev{
\subsection{Dealing with Long Text Streams}
}

\minrev{
The proposed model explicitly memorizes the information about entities and relations of the current and past stories into distinct (i.e., independent) memory locations (Section \ref{sec:cg}). This strongly alleviates the catastrophic forgetting problem \cite{mccloskey1989catastrophic}, that is usually due to memory interference among multiple entities or relations (e.g.,  due to weight sharing). 
The experimental analysis of Section \ref{subsec:results}, where the same stream is read a second time (``Re-Reading''), suggests that the proposed system does not incur into serious catastrophic forgetting issues, since its recognition accuracy  improves once reading again the text stream, without additional supervisions (Figure \ref{fig:all_last_E_R}).
Moreover, even if the encoding module of Section \ref{sec:enc} is based on RNNs, we are not exposed to the problem of learning long-term dependencies with such RNNs \cite{bengio1994learning}. As a matter of fact, the RNNs are only responsible of encoding each mention and its context within a single sentence, whose usual length is efficiently manageable with LSTMs without any long-term dependency problems.  
}


\minrev{
We performed a further experimental analysis aimed at evaluating the memorization skills of the system when dealing with long text streams. In particular, we provided the system an incremental number of stories, and we periodically evaluated its capability of correctly recognizing the entities of the first story of the stream. As long as the system reads more data, this procedure provides a clear indication of the persistence of the system memory.
However, the system is still exposed to ambiguity issues, since multiple instances might share the same mention. For this reason, we considered different experimental settings with different levels of ambiguity. 
%
}

\minrev{
In detail, we considered four scenarios in the Simple Story Dataset, referred to as $A$, $B$, $C$, $D$, respectively. 
}
\minrev{
In each of them, we generated $10$ different long streams ($\approx 2600$ sentences in each stream), randomly shuffling the involved stories.
For each stream, we selected the first story as \textit{target} story. The model reads the stream and learns in an online manner, receiving supervisions about the mentioned entities, and constantly adapting its internal parameters as in the experiments of Section \ref{subsec:learning_setting}. 
After having processed $t$ stories, we temporarily freeze the whole system, and we process  the target story again, measuring the prediction accuracy on the entities of this story. By increasing the number $t$ of intermediate stories, we evaluate how the systems keeps memory of the target story as long as the stream grows.
}

\minrev{
The four settings differ in how the stream has been constructed, and they introduce an increasing level of ambiguity.
In the first setting ($A$), intermediate stories in the stream neither contain entities nor mentions belonging to the target story. Reading the stream does not introduce any ambiguity in the contents of the target story. This allows us to evaluate whether the system is forgetting information as long as time passes, completely discarding ambiguity issues. 
In setting $B$ we include pronouns in the intermediate stories. Mentions related to pronouns are obviously very ambiguous, hence the model will have to rely on the temporal hypothesis (Section \ref{sec:cg}) to be able to disambiguate them. 
The intermediate stories of setting $C$ can contain mentions that are also used in the target story, and these mentions may or may not refer to the same entities of the target story. 
In this case, the system must not only remember what was read in the target story, but it must also gain stronger skills in disambiguating entities.
Finally, in the last scenario ($D$), we ensure that the intermediate stories never refer to the entities of the target story. However, we still allow the stream to use mentions that are shared with the ones that are used in the target story
, simulating a strong data drift.
}

\minrev{
We explicitly computed the degree of ambiguity of each stream for all the settings (referred to as $Amb_A$, $Amb_B$, $Amb_C$, $Amb_D$), that is the average number of entities associated to each mention of the target story up to step $t$. Table \ref{tab:long_streams} reports both the degrees of ambiguity and the recognition accuracy of the entities of the target story (averaged over the $10$ generated streams). We report the results after $t=0$, $10$, $50$, $100$, $150$ intermediate stories ($0$, $180$, $880$, $1700$, $2600$ intermediate sentences, respectively).
}
\minrev{
}
\begin{table}
	\centering
	\caption{\minrev{Average recognition accuracy (on 10 different streams) of the entities of target stories, with an increasing number $t$ of intermediate stories for settings A, B, C, D ($Acc_A$, $Acc_B$, $Acc_C$, $Acc_D$), and average number of entities per mention (degree of ambiguity $Amb_A$, $Amb_B$, $Amb_C$, $Amb_D$).}}
    \begin{tabular}{llllll}
    \toprule
    & $t=0 $& $t=10$     & $t=50$   & $t=100$   & $t=150$  \\
    \midrule
    $Amb_A$ & 1.03 & 1.03 & 1.03 & 1.03 & 1.03\\
    $Amb_B$ & 1.05 & 1.42 & 2.14 & 2.42 & 2.65\\
    $Amb_C$ & 1.04 & 1.50 & 2.49 & 3.28 & 3.64\\
    $Amb_D$ & 1.03 & 1.83& 3.48 & 4.54& 5.06 \\
    \midrule
    $Acc_A$ &  98.41 & 98.41  & 98.41 & 98.41 & 98.41           \\ 
    $Acc_B$ & 99.62  & 96.88  & 92.03 & 94.03 & 94.03          \\ 
    $Acc_C$ & 98.89 &  85.05     & 72.25      &    70.30  & 71.85           \\ 

    $Acc_D$ & 97.98 & 85.83 & 71.32 & 65.93 & 63.25  \\
 \bottomrule
    \end{tabular}
    \label{tab:long_streams}
\end{table}
\minrev{
All the four settings show an accuracy close to $100\%$ at $t=0$, as expected, where the few errors are mostly due to pronouns, since the system has not developed strong co-reference skills, as expected.
For $t>0$, results in setting $A$ confirm that the model does not forget past information when processing text streams that do not interfere with it.
Setting $B$ shows an interesting behaviour, in which accuracy is reduced as long as $t$ grows up to $50$, and then it increases again. This indicates that while at the beginning the system misclassifies pronouns in the target story, it is able to improve pronoun resolution while reading several stories, thus adjusting the temporal hypothesis of Equation \ref{eq:tp}.
In both settings $C$ and $D$, as expected, results indicate that performances degrade as the number of intermediate stories increases, due to the increasing number of ambiguous mentions. The model clearly performs better in setting $C$ than in setting $D$, remarking the positive effects of refining the disambiguation units while reading information about the entities of the target story in multiple occasions. Finally, it is worth noticing that the degradation of the performances is directly related to the degree of ambiguity in the streams. This is a clear hint that long streams increase the difficulty of the task not due to their length but to their intrinsic ambiguity.
}

%% file: conclusions.tex
We presented an end-to-end model to process text streams, where mentions to entities and relations are \rev{detected}, disambiguated, and eventually added to an internal KB, that, differently from many existing works, is not-given-in-advance.
Our model is capable of performing one-shot learning, self-learning, and it learns to resolve co-references. It has shown strong disambiguation and discovery skills when tested on a stream of sentences organized into small stories (\rev{we also created a new dataset that we publicly made available for further studies}), even when a few, sparse supervisions are provided. We also showed how it can improve its skills by continuously reading text. 
Our future work will focus on exploiting entities and relations structured into facts, higher-level reasoning, types, dynamic re-organization of the KB.